\documentclass[journal,lettersize]{IEEEtran}
\usepackage{mmstyles}
\usepackage{graphicx}
\usepackage{comment}
\usepackage{hyperref}
\usepackage{microtype}
\usepackage{subcaption}
\usepackage{multirow}
\usepackage{color}

\usepackage[T1]{fontenc}

\ifCLASSINFOpdf
\else
\fi

\usepackage{amsmath}
\usepackage[cmintegrals]{newtxmath}

\begin{document}
%
\title{Texture Memory-Augmented \\ Deep Patch-Based Image Inpainting}

\author{Rui~Xu, Minghao~Guo, Jiaqi~Wang, Xiaoxiao~Li, Bolei~Zhou and Chen~Change~Loy}

\twocolumn[{
	\renewcommand\twocolumn[1][]{#1}
	\maketitle
	\begin{center}
		\centering
		\includegraphics[width=\linewidth]{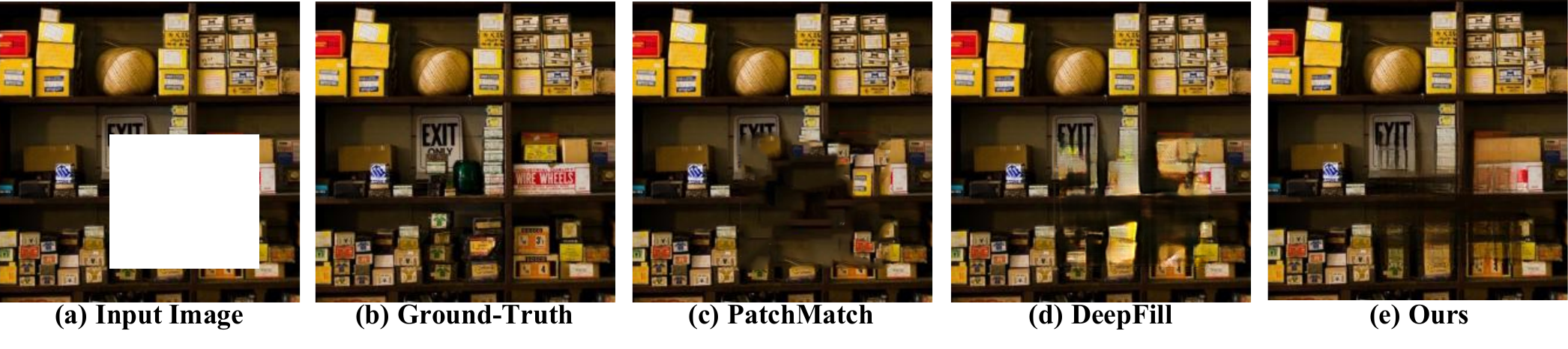}
		\vskip -0.2cm
		\captionof{figure}{Comparing inpainting results generated by (c) a patch-based approach, (d) a deep learning-based method, and (e) the proposed Texture Memory Augmented approach. Our method is capable of recovering both the global semantic structure and local texture details, taking the best of both worlds of patch-based paradigm and deep learning-based method.}
		\label{fig:teaser}
	\end{center}
}]

\begin{abstract}
Patch-based methods and deep networks have been employed to tackle image inpainting problem, with their own strengths and weaknesses.
Patch-based methods are capable of restoring a missing region with high-quality texture through searching nearest neighbor patches from the unmasked regions. However, these methods bring problematic contents when recovering large missing regions.
Deep networks, on the other hand, show promising results in completing large regions. Nonetheless, the results often lack faithful and sharp details that resemble the surrounding area.
By bringing together the best of both paradigms, we propose a new deep inpainting framework where texture generation is guided by a texture memory of patch samples extracted from unmasked regions.
The framework has a novel design that allows texture memory retrieval to be trained end-to-end with the deep inpainting network. In addition, we introduce a patch distribution loss to encourage high-quality patch synthesis.
The proposed method shows superior performance both qualitatively and quantitatively on three challenging image benchmarks, i.e., Places, CelebA-HQ, and Paris Street-View datasets.\footnote{Code will be made publicly available in https://github.com/open-mmlab/mmediting.}
\vspace{-0.25cm}

\end{abstract}


%
\IEEEpeerreviewmaketitle

\section{Introduction}
\label{sec:intro}

Image inpainting aims at filling in missing regions of a given image with consistent and coherent contents. 
The challenge of image inpainting lies in two aspects, \ie, reconstructing the missing global structure and synthesizing realistic local textures coherent to unmasked regions. 

Image inpainting is a long-standing problem in the field of computer vision.
Classic patch-based methods~\cite{barnes2009patchmatch,he2012statistics} perform patch matching within an image and fill the missing region that plausibly matches the remaining image content.
While textures are faithfully recovered in many cases, patch-based methods often fail to restore the global geometric structure, especially when presented with corrupted images with large missing regions. A failure case from the popular PatchMatch method \cite{barnes2009patchmatch} is shown in Fig.~\ref{fig:teaser}(c). 
A significant progress in image inpainting has been attained through the usage of deep convolutional neural networks (CNN) and generative adversarial network (GAN)~\cite{liu2018image,yu2018free,yu2018generative}.
Deep inpainting methods formulate image inpainting as a conditional image generation task.
A convolutional encoder-decoder is typically trained with an adversarial loss to restore reasonable structures and textures.
These methods demonstrate a promising performance in hallucinating the missing contents with coherent semantic structures.
However, given the high complexity of scenes and textures in the wild, it is still challenging for a deep generative model to inpaint arbitrary images. 
%
%

The goal of this work is to bring together the best of classic patch-based and deep learning based approaches, \ie, leveraging deep networks to recover the global structure of the missing region and exploiting the notion of patch matching to restore the details.
It is non-trivial to combine the two vastly different paradigms in a unified end-to-end model. 
Specifically, the learning process of classic patch-based methods is not end-to-end, requiring expensive patch-synthesis based on optimization \wrt the content and texture constraints. 
A plausible solution is to perform implicit patch selection through contextual attention, which has been widely used in deep inpainting literature~\cite{yu2018free,yu2018generative,ren2019structureflow,yi2020contextual} to extract information from unmasked regions.  Contextual attention module adopts a softmax function to fuse all candidate texture features for generation. In our experiments, we observe that interpolation in texture space may lead to blurry output and unnatural artifacts, as shown in Fig.~\ref{fig:teaser}(d).

To mitigate the aforementioned challenges, we formulate a new mechanism for patch matching, retrieval, and generation within the deep network framework. The proposed method, which we call as \textit{Texture Memory-Augmented Deep Patch-Based Image Inpainting} (T-MAD) has a few appealing properties.
Differing to implicit patch selection in contextual attention which involves interpolation in texture space, we present a new sampling strategy that provides a more explicit guidance towards the generation of patch-level details.
This is achieved through the notion of \textit{texture memory}, which contains patches extracted from unmasked regions of the input image.
With a tailored design, the patch matching and retrieval steps in texture memory is end-to-end trainable.
Guided by the coarse result (which can be generated by any existing deep inpainting networks), patches with high similarity to the coarse result are retrieved from the memory, and their features are then used to guide the recovery of patch-level textures on top of the coarse global reconstruction.
With this retrieve-and-guidance design, we pay more attention to the generation of patch-level details, the results  are thus less susceptible to blurry and unnatural artifacts. An example is shown in Fig.~\ref{fig:teaser}(e), more examples can be found in the experiments section. 

The main contribution of this paper is an effective image inpaiting method that unifies patch- and deep learning-based approaches. We devise an effective method that enables back-propagation for patch matching and retrieval from texture memory. Hence, these steps can be learned in an end-to-end manner within our unified framework. 
We also propose an effective adversarial loss at patch level to better capture patch statistics. Different from existing adversarial losses for learning the distribution of the whole dataset, we adopt patch distribution loss to model the texture distribution of a target image. 

We show the advantages of retrieve-and-guidance framework on various benchmark data with random rectangle masks and free-form irregular holes. In particular, our method outperforms state-of-the-art methods CRA~\cite{yi2020contextual} in two different mask settings quantitatively and qualitatively.
We further show that the notion of texture memory can be easily adapted into other pipelines, \eg, DeepFill~\cite{yu2018free}, as a useful post-processing module to improve the quality of image inpainting.

\section{Related Work}
\label{sec:related_work}

\noindent\textbf{Patch-Based Image Inpainting.}
The seminal work PatchMatch~\cite{barnes2009patchmatch} and its successful variants~\cite{he2012statistics,yang2017high,lotan2016needle} have been widely used in image editing and inpainting.
These optimization-based methods regard image inpainting as finding the optimal patches for each masked location with manually designed constraints.
These methods have shown to be robust to the variance of textures and input resolution. 
However, the optimization process is expensive.
In addition, these methods are inherently limited by the ability of generating novel contents and they tend to fail to preserve a reasonable global structure when the missing region is large.
Yang \etal~\cite{yang2017high} adopt a deep model to reconstruct the coarse content and apply VGG network~\cite{simonyan2014very} to compute a perceptual loss as an extra constraint in the patch matching paradigm. 
Nevertheless, their method requires expensive multi-scale patch synthesis based on optimization \wrt the content and
texture constraints. In addition, the learning of coarse generation and texture synthesis cannot be trained end-to-end.

\begin{figure*}[t]
    \centering
    \includegraphics[width=0.97\linewidth]{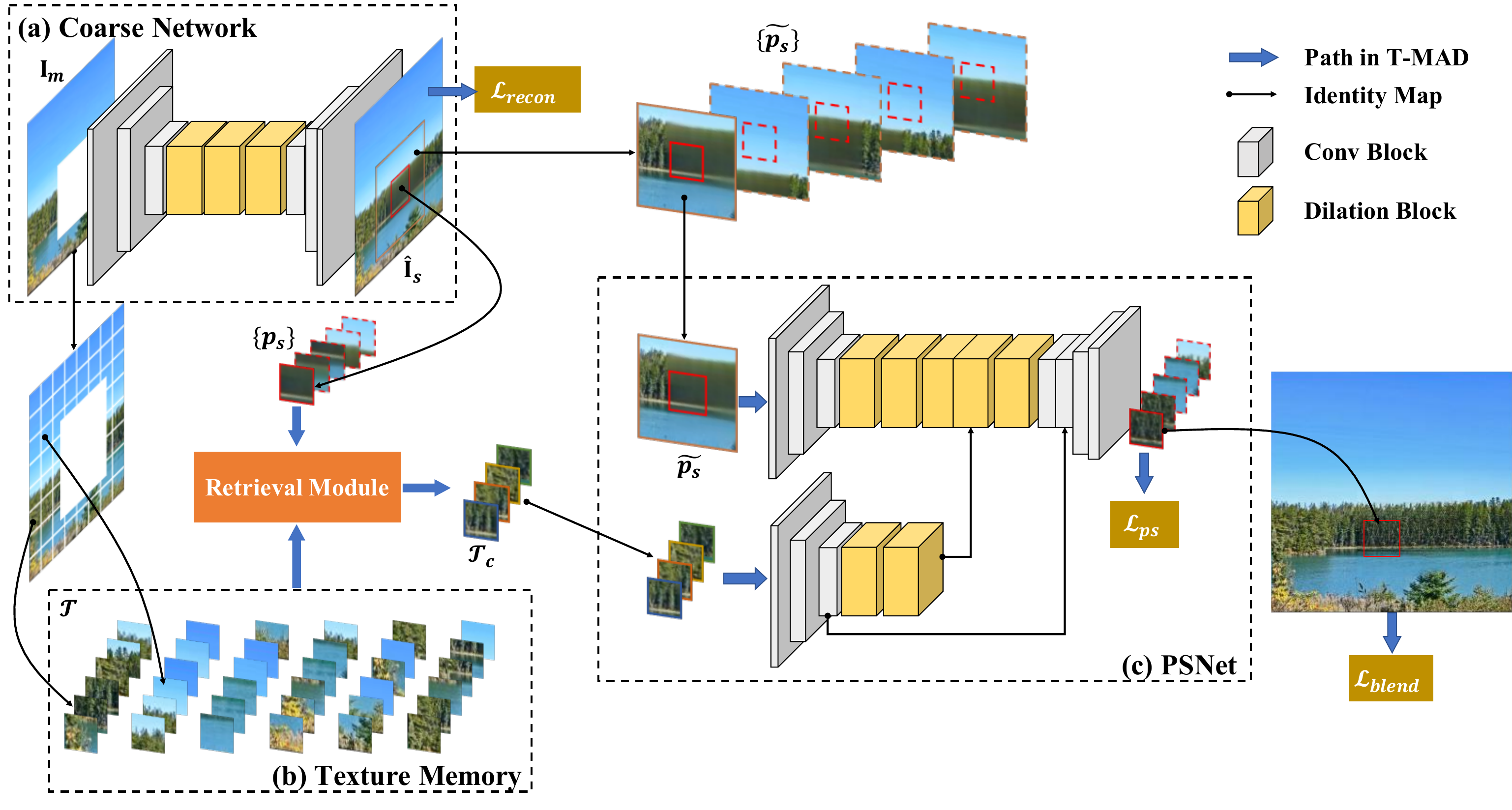}
    \caption{Framework of the proposed T-MAD. There are three key modules: (a) A coarse network for generating the coarse global results, (b) A texture memory for maintaining texture priors from unmasked regions, and (c) A patch synthesis module (PSNet) for patch synthesis, conditioned on the coarse global results and patches retrieved from the texture memory. Patch retrieval from the texture memory is formulated as an end-to-end trainable module.}
    \label{fig:framework}
    \vspace{-0.35cm}
\end{figure*}

\noindent\textbf{Deep Image Inpainting.}
Existing deep inpainting methods fall into two categories, single-stage and multi-stage approaches.
Single-stage approaches~\cite{iizuka2017globally,liu2018image,zeng2019learning,hong2019deep,wang2021dynamic,li2020recurrent} adopt an encoder-decoder network with multiple losses to recover the corrupted region directly. 
The networks are trained to jointly capture the structure and texture information in a single pass. 
As for multi-stage approaches, existing models can be further divided into two subcategories. 
Methods from the first category~\cite{yu2018free,yu2018generative,zhu2021image,zeng2020high} adopt a coarse-to-fine framework where synthesis is gradually refined.
Approaches in the second category reconstruct structural information in the first stage as a prior to guide the following stages for synthesizing detailed textures~\cite{nazeri2019edgeconnect,xiong2019foreground,ren2019structureflow,liao2020guidance}.

Although the aforementioned methods are capable of generating better global structures compared with the traditional optimization-based methods, the generated textures are still unsatisfactory for two reasons.
First, they require the model to synthesize detailed textures on the entire image. The requirement inevitably adds difficulty in training a high-resolution GAN.
Second, these methods typically encounter difficulty in generating complex textures for in-the-wild images since such textures of the target image may not be seen during training. 
To address these problems, we synthesize textures at patch level and leverage the texture memory constructed from unmasked regions. 

The fact that our method retrieves plausible patches from a texture memory can be loosely regarded as a kind of attention. Attention mechanism~\cite{vaswani2017attention,wang2018non,devlin2018bert} has been widely used in deep inpainting literature~\cite{yu2018free,yu2018generative,ren2019structureflow,yi2020contextual} to extract information from valid regions. The key differences between our method and the popular contextual attention approach lie in the ultimate generative goal and patch sampling strategy. 
First, T-MAD aims at synthesizing high-quality patches with guidance of texture memory to inpaint the hole while contextual attention approaches~\cite{yu2018free,yu2018generative,ren2019structureflow,yi2020contextual} directly inpaint at image level. Second, while contextual attention module adopts softmax function to fuse candidate texture features for generation, the proposed T-MAD uses a differentiable patch retrieval module to sample similar patches to guide the generation process and thus avoids interpolation in texture space. The differences lead to the better capability in T-MAD in generating visually pleasing details.

\section{Methodology}
\label{sec:approach}


Unlike previous deep models that perform inpainting on an entire image, our T-MAD approach focuses on the corrupted regions to synthesize high-quality patches. 
These patches are then tiled into the missing region accordingly. As illustrated in Fig.~\ref{fig:framework}, our T-MAD contains three modules. 
First, as shown in Fig.~\ref{fig:framework}(a), we roughly estimate the missing contents with a coarse network trained with $l_1$ reconstruction loss, producing a reasonable global structural prior. 
Next, we split the coarse result into non-overlapping patches of equal size, $\{p_s\}$, and further prepare $\tilde{p}_s$ that are extracted from the coarse result according to $\{p_s\}$ with a larger neighboring context.
Meanwhile, as shown in Fig.~\ref{fig:framework}(b), we maintain a texture memory containing multiple patches extracted from the unmasked region.
%
Guided by the coarse result patches, the retrieval module selects the most similar texture patches as a prior for subsequent patch synthesis step in patch synthesis module (PSNet), as depicted in Fig.~\ref{fig:framework}(c). 
By fusing information from both the coarse structure and the patches with native texture, the proposed patch synthesis module generates high-quality textures that are coherent with the unmasked region and meet the global structure constraint.
%

The section is organized as follows: Sec.~\ref{sec:cnet} introduces the proposed coarse network for recovering rough structure information. The construction and retrieval procedure of the texture memory are clarified in Sec.~\ref{sec:pool}. Finally, the patch synthesis module is explained in Sec.~\ref{sec:psmodule}

\subsection{Coarse Network for Structural Prior}
\label{sec:cnet}

A coarse network, as depicted in Fig.~\ref{fig:framework}(a), is used to generate a coarse and global estimation, of which the content will be used to guide the subsequent patch synthesis process.
One can adopt any existing deep inpainting network for this purpose.
In this work, the network takes an encoder-decoder structure, accepting a masked image $\mathbf{I}_m$ and the corresponding binary indicating mask $\mathbf{M}$ as input. 
The coarse network produces an initial coarse result $\mathbf{I}_s$ with the same size as the input.
The completed image $\hat{\mathbf{I}}_s$ can be obtained by combining $\mathbf{I}_s$ with valid pixels from the unmasked region.
%

A large receptive field is desirable for recovering missing structural information.
Adopting dilated residual blocks~\cite{chen2017deeplab,nazeri2019edgeconnect} is a feasible option to enlarge the receptive field. We opt to adopt eight residual blocks in our coarse network, after taking into account both the model capacity and the training cost.
%
%
%
Meanwhile, we substitute all up-sampling modules with CARAFE~\cite{wang2019carafe} to allow content-aware feature reassembly for better structural reconstruction.
The coarse network is supervised with $l_1$ reconstruction loss between the output and the original image $\mathbf{I}$
\begin{equation}
    \mathcal{L}_{\rm{recon}}=\lambda^s_{m} \left\|\mathbf{I}_s - \mathbf{I} \right\|_1 \odot \mathbf{M} +  \lambda^s_{v} \left\|\mathbf{I}_s - \mathbf{I} \right\|_1 \odot (1 -\mathbf{M}) ,
    \label{eq:recon-loss}
\end{equation}
where $\odot$ denotes an element-wise product.
We empirically find that setting $\lambda_m^s=5$ and $\lambda_v^s = 0$ brings better results for random rectangle missing regions, while $\lambda_m^s=6$, and $\lambda_v^s=1$ is a better choice for irregular holes.
After obtaining $\hat{\mathbf{I}}_s$, we apply a $k_{p}\times k_{p}$ sliding window to cut the completed region in $\hat{\mathbf{I}}_s$ into multiple non-overlapping patches $\{p_s\}$. 
The patches $\{p_s\}$ that encapsulate the initial structure prior will guide the subsequent texture retrieval and patch synthesis process.

\subsection{Retrieving Visual Prior from Texture Memory}
\label{sec:pool}
\noindent\textbf{Texture Memory.} 
%
Different from existing deep learning approaches, T-MAD has a carefully designed texture memory containing texture patches extracted from the unmasked region. Together with the coarse global results, these patches will be used to guide the patch synthesis process.

Similar to the procedure in extracting coarse result patches, as shown in Fig.~\ref{fig:framework}(b), a $k_p\times k_p$ sliding window is adopted to extract valid texture patches from $\mathbf{I}_m$. 
We perform a dense sampling to keep sufficient texture patches in our memory pool.
A reasonable choice of sliding window stride is $s_{pool}=k_p / 2$.
Given the texture patches, we randomly select $\mathbb{N}_{pool}$ patches to construct the final texture pool $p_t \in \mathcal{T}$.
As for irregular holes like~\cite{yu2018free}, $s_{pool}$ is set to $k_p / 4$ and we choose texture patches constaining the least masked regions as $p_t$ for further increasing the number of valid patches.
In our implementation, $\mathbb{N}_{pool}$ is set to 100. For the patch size in the texture memory, we recommend a size of $k_p=32$. 
This setting gives a good balance between the final texture quality and the training cost.
It is noteworthy that unlike previous methods~\cite{huang2016temporally} in computer graphics, we do not need to introduce complex distortions to the texture patches since these patches mainly serve as a condition for the following patch synthesis module (discussed in Section~\ref{sec:psmodule}). 
The module is capable of transferring the texture effectively onto the inpainted image.



The selected patches in the texture memory are retrieved to guide the process of patch synthesis.
The retrieval module consists of two components to perform \emph{correspondence embedding} and \emph{differentiable sampling}, respectively. 

\vspace{0.1cm}
\noindent\textbf{Correspondence Embedding.}
The correspondence embedding component constructs a similarity matrix $c$ between the coarse patches $p_s^i$ and the texture patches in the texture memory $p_t \in \mathcal{T}$. A non-local module~\cite{wang2018non} are adopted to find such correspondance:
%
\begin{equation}
    c(p_s^i, p_t^j)=\theta(p_s^i)^{\scriptscriptstyle\mathsf{T}} \phi(p_t^j),
    \label{eq:non-local}
\end{equation}
where $\theta$ and $\phi$ are two shallow convolutional networks.
As the scale of $||\phi(x_j)||$ introduces an unreasonable norm term to $c(p_s^i, p_t^j)$, we reformulate the original equation as:
\begin{equation}
    c(p_s^i, p_t^j)=\theta(p_s^i)^{\scriptscriptstyle\mathsf{T}} \frac{\phi(p_t^j)}{||\phi(p_t^j)||} .
    \label{eq:nonlocal-mod}
\end{equation}
The normalization of $\phi(p_t^j)$ removes the influence of the norm value on learning such specific similarity. 
The output of the non-local module is normalized by a softmax function to establish the correspondence $\mathcal{S}$ between every patch in $\{p_s\}$ and in $\mathcal{T}$. 

\vspace{0.1cm}
\noindent
\noindent\textbf{Differentiable Sampling.}
Previous works with memory pool~\cite{voigtlaender2019feelvos,wu2019long} predict softmax value to form the pool in a weighted-sum manner. 
However, an interpolation of different textures prevents one from generating meaningful textures as the weighted sum would result in blurry outputs.
In our approach, for each patch $p_s$, we sample $\mathbb{N}_c$ texture patches $p_c\in\mathcal{T}_c$ from $\mathcal{T}$ as candidates. These patches are regarded as those that are most similar to the coarse patch $p_s$, yet contain diverse and rich priors on textures.
In Fig.~\ref{fig:framework}, we use an example patch with solid red bounding box to show the retrieval procedure.

It is known that the sampling operation prevents gradients from having a direct path in back-propagation.
Inspired by Binary Network~\cite{courbariaux2016binarized}, we formulate a discrete weighted sum to ensure that the output in the forward path is the exact patch we need. Meanwhile, a soft value will be adopted in the back-propagation.
Suppose the obtained similarity vector is $\mathcal{S}\in \mathbb{R}^{1\times \mathbb{N}_{pool}}$.
To sample the $i$-th patch from $\mathcal{T}$, we first construct an indicating vector $\mathbf{1}_i\in \mathbb{R}^{1\times\mathbb{N}_{pool}}$, in which only the $i$-th index is 1 and the others are 0.
The $i$-th index in $\mathbf{1}_i$ indicates the patches with top scores or ranks.
We can obtain the exact $i$-th patch $p_i^s$ from $\mathcal{T}$ through Eq.~\eqref{eq:sample}, where $\otimes$ denotes a weighted-sum operation and \texttt{detach} indicates variables that do not require gradient in the back-propagation process.
%
%
\begin{equation}
        p_c^i = \mathbf{1}_i\otimes\mathcal{T} 
              = \{(\mathbf{1}_i - \mathcal{S}).\texttt{detach} + \mathcal{S}\} \otimes\mathcal{T} .
    \label{eq:sample}
\end{equation}
The back-propagation through our sampling function is as 
\begin{equation}
    \label{eq:sample-bp}
    \begin{split}
        \frac{\partial p_c^i}{\partial w} &=\frac{\partial \mathbf{1}_i}{\partial w}\otimes\mathcal{T} \\
                                          &=(\frac{\partial(\mathbf{1}_i - \mathcal{S}).\texttt{detach}}{\partial w} + \frac{\partial \mathcal{S}}{\partial w}) \otimes\mathcal{T}\\
                                          &=(\mathbf{0} + \frac{\partial \mathcal{S}}{\partial w})\otimes\mathcal{T}=\frac{\partial \mathcal{S}}{\partial w}\otimes\mathcal{T},
    \end{split}
\end{equation}
where $w$ indicates learnable parameters in this module.
Due to the \texttt{detach} operation, the back-propagation procedure regards $\mathbf{1}_i - \mathcal{S}$ as a constant value so that gradients can be propagated to $\mathcal{S}$ successfully.
This reformulation allows the proposed T-MAD to be trained in an end-to-end manner.

Gumbel-Softmax~\cite{jang2016categorical} replaces non-differentiable sampling from a categorical distribution with a differentiable sampling from a novel Gumbel-Softmax distribution. However, Gumbel-Softmax can easily sample the one with the highest score but cannot sample other candidates with specific scores or ranks.

\subsection{Patch Synthesis Module -- PSNet}
\label{sec:psmodule}
In this module, we focus on generating high-quality patches guided by coarse result patch $\{p_s\}$ and the selected texture prior $\mathcal{T}_c$.
The goal is not only to synthesize textures without blurry artifacts but also to match the texture distribution of the unmasked region.
Importantly, $\{p_s\}$ captures global structural prior, which guides the texture synthesis module to have a reasonable global semantic structure.
Through synthesizing textures on each small patch, we expect to lift the burden of training a general conditional GAN on the entire image.

\noindent\textbf{Network Architecture.}
As shown in Fig.~\ref{fig:framework}(c), an encoder-decoder network is adopted as a backbone for texture synthesis, while another simple encoder is designed to encode the texture prior $\mathcal{T}_c$ for PSNet. 
Two skip connections provide the backbone encoder-decoder network with pyramid texture information from the texture memory.
The pyramid texture priors from $\mathcal{T}_c$ are concatenated to the intermediate feature maps in the backbone encoder-decoder.

The inputs to the backbone encoder-decoder and texture prior encoder are different.
To preserve the consistency between the synthesized texture and its neighboring patches, the backbone encoder-decoder takes the initial coarse patches $\{p_s\}$ and its neighboring patches in $\hat{I}_s$ as inputs.
This can be achieved by using a larger sliding window $3k_p\times 3k_p$ to extract larger patches $\tilde{p}_s$ with stride $s_{\tilde{p}_s}=k_p$. 
%
%
As for the input for the texture encoder network, all patches in $\mathcal{T}_c$ will be concatenated as the input.

A notable feature of our method is that it can perform patch synthesis in parallel. For parallel computation, all $\tilde{p}_s^i$ and the corresponding $p_c^i \in \mathcal{T}_c$ are reshaped into a single batch dimension.
Compared with existing deep image inpainting models~\cite{iizuka2017globally,yu2018free} that need to process an entire image, our approach consumes 20\% less memory in training and inference by processing non-overlapped patches of missing regions in parallel.

\noindent\textbf{Patch Distribution Loss.}
%
To leverage the texture prior and match the texture distribution across observed patches, we propose a patch-level adversarial loss. 
%
%
%

%
%
In our task, besides the ground-truth patches $p_{gt}$, the synthesized patches should also subject to a similar distribution with candidate texture pool $\mathcal{T}_c$.
Following the notion of adding relativistic comparison between fake results and real samples in RAGAN~\cite{jolicoeur2018relativistic}, we introduce the comparison at patch level into adversarial training.
The detailed formulation is shown in Eq.~\eqref{eq:psgan-d} and Eq.~\eqref{eq:psgan-g}
\begin{equation}
    \begin{split}
    \mathcal{L}_D &=\mathbb{E}_{x_r\sim \mathbb{P}(\mathcal{T}_c, p_{gt})}[f_1(C(x_r)-\mathbb{E}_{x_f\sim Q}C(x_f))] \\
    &+\mathbb{E}_{x_f\sim Q}[f_2(C(x_f)-\mathbb{E}_{x_r\sim \mathbb{P}(\mathcal{T}_c, p_{gt})}C(x_r))],
    \end{split}
    \label{eq:psgan-d}
\end{equation}
\begin{equation}
    \begin{split}
        \mathcal{L}_G &=\mathbb{E}_{x_r\sim \mathbb{P}(\mathcal{T})}[g_1(C(x_r)-\mathbb{E}_{x_f\sim Q}C(x_f))] \\
         &+\mathbb{E}_{x_f\sim Q}[g_2(C(x_f)-\mathbb{E}_{x_r\sim \mathbb{P}(\mathcal{T}_c, p_{gt})}C(x_r))],
    \end{split}
    \label{eq:psgan-g}
\end{equation}
\hspace{-0.12cm}where $f_1$, $f_2$, $g_1$, $g_2$ are scalar-to-scalar functions, $\mathbb{P}(\cdot)$ indicates the distribution of certain data, $C(x)$ is the non-transformed discriminator output and $\mathcal{T}$ is our texture memory containing diverse texture patches from unmasked regions.
Here, we provide the discriminator $D$ with $(\mathcal{T}_c, p_{gt})$ to encourage the discriminator to predict the difference between $x_f$ and the most similar textures. 
Meanwhile, in the generator $G$, a compensation term is added by computing the distance between $\mathbb{P}(\mathcal{T})$ and the mean distribution of $\{x_f\}$ in case of failing to find the suitable patches $\{p_c\}$.
In addition, this compensation term encourages the mean style of the synthesized patches to be coherent with unmasked regions.

To encourage the patch synthesis module to generate patches following the constraints encoded in the coarse patches $\{p_s\}$, we also adopt a $l_1$ loss $\mathcal{L}_{l_1}$ as another supervision. 
Moreover, a perceptual loss $\mathcal{L}_{percep}$ is also adopted to further improve the perceptual quality.
In $\mathcal{L}_{percep}$, we apply the officially pre-trained VGG-19 as the inception network to extract features.
Unlike previous works~\cite{liu2018image,nazeri2019edgeconnect} that employ features after pooling layers, we adopt features before ReLU function to reduce artifacts.
%
%
We use features from `conv\_5', `conv\_9' and `conv\_15' before ReLU in our perceptual loss.
%
The total loss for patch synthesis is:
\begin{equation}
    \mathcal{L}_{\rm{ps}} = \lambda_{\rm{gan}}^{pd}\mathcal{L}_{\rm{gan}}^{pd} + \lambda_{l_1}\mathcal{L}_{l_1} + \lambda_{\rm{percep}}\mathcal{L}_{\rm{percep}},
    \label{eq:psloss}
\end{equation}
where $\mathcal{L}_{\rm{gan}}^{pd}$ only contains the generator part in Eq.~\eqref{eq:psgan-g}.

The proposed patch-level adversarial loss is different from PatchGAN~\cite{CycleGAN2017, isola2017image} or patch discriminator~\cite{demir2018patch,shaham2019singan}, which is widely adopted in recent deep inpainting methods~\cite{nazeri2019edgeconnect,yu2018free, demir2018patch}. 
These approaches encourage the discriminator to determine the quality of a synthesized image by classifying if each $N\times N$ patch in the image is real or fake. The scores of all patches are averaged to get the final score of the image as the loss.
%
%
The main differences between PatchGAN and our approach are: 1) we model the distribution at patch level rather than image level, and 2) the goal of our patch synthesis module is to match the distribution of unmasked regions while PatchGAN still focuses on synthesizing an image \wrt the entire dataset distribution.

\noindent\textbf{Blending Loss.}
To generate the final image $\hat{\mathbf{I}}_p$, the synthesized patches will be allocated back to the original position of the input to fill in the missing region.
To remove the boundary artifacts and preserve the consistency among neighboring patches, we apply a total variation loss.
A global patch discriminator~\cite{CycleGAN2017} on $\hat{\mathbf{I}}_p$ is also helpful in this step. The blending loss is as
\begin{equation}
    \mathcal{L}_{\rm{blend}} = \lambda_{\rm{gan}}^{gl}\mathcal{L}_{\rm{gan}}^{gl} + \lambda_{\rm{tv}}\mathcal{L}_{\rm{tv}},
    \label{eq:blendloss}
\end{equation}
where $\mathcal{L}_{\rm{tv}}$ is a widely-used regularization~\cite{liu2018image} to smooth the generated contents. In the detailed implementation, $\mathcal{L}_{\rm{tv}}$ is calculated by averaging the differences between neighboring pixels.
The complete loss function of T-MAD contains the three terms provided in Eq.~\eqref{eq:recon-loss}, Eq.~\eqref{eq:psloss} and Eq.~\eqref{eq:blendloss}, corresponding to the loss of coarse network, patch synthesis and the final tiling process.
\begin{equation}
    \begin{split}
    \mathcal{L}_{\rm{total}}= \mathcal{L}_{\rm{recon}} + \mathcal{L}_{\rm{ps}} + \mathcal{L}_{\rm{blend}}.
    \end{split}
    \label{eq:totalloss}
\end{equation}

\section{Experiments}
\label{sec:experiments}

\noindent
\textbf{Implementation Details.}
We follow the design of convolution blocks in PGGAN~\cite{karras2017progressive} to prepare our encoders and decoders as it proves to be effective on image generation. For all encoders in our T-MAD, a general downsampling block with `conv$3\times 3$, conv$3\times 3$, Downsample' is adopted and we halve the image resolution using nearest neighbor filtering. 
As for decoders in our framework, we apply `Upsample, conv$3\times 3$, conv$3\times 3$' as the basic building unit, named as upsampling block. The content-aware reassembly method, CARAFE~\cite{wang2019carafe}, is chosen as our `Upsample' method. 
In Tab.~\ref{tab:block}, we show the detailed number of the basic blocks in different modules. Note that we just apply a standard residual block~\cite{nazeri2019edgeconnect} in PSNet without any dilation operation. This is because PSNet aims at synthesizing high-quality textures at patch level and large receptive field is not necessary. In addition, we use leaky ReLU with leakiness 0.2 in all layers of the whole networks, except for the last layer in the decoder that uses linear activation.  We do not apply batch normalization or instance normalization in our T-MAD. In patch retrieval module, only two convolution layers with `conv$3\times 3$, conv$1\times 1$' are adopted for efficient feature extraction.

\begin{table}[t]
    \centering
    \caption{The number of basic blocks in different modules. `input-conv' and `out-conv' contain a $1\times 1$ convolution layer for channel transformation. `PSNet-FeatEnc' denotes the encoder in PSNet for extracting features from $\mathcal{T}_c$. The dilated residual block is denoted as `res-block'.}
    \vspace{-5pt}
    \begin{tabular}{c|c|c|c}
                & Coarse Net & PSNet-Backbone & PSNet-FeatEnc \\ \hline \hline
    input-conv  & 1              & 1              & 1             \\ \hline
    down-block  & 3              & 2              & 2             \\ \hline
    res-block   & 8              & 5              & 2             \\ \hline
    up-block    & 3              & 3              & 0             \\ \hline
    out-conv & 1              & 1              & 0            
    \end{tabular}
    \vspace{-8pt}
    \label{tab:block}
\end{table}

\begin{table}[tb]
    \centering
    \caption{Hyper-parameters in our T-MAD framework for random rectangle holes and free-form holes.}
    \vspace{-5pt}
    \label{tab:param}
    \small
    \begin{tabular}{c|c|c}
        \multicolumn{2}{c|}{}                            & Hyper-Parameters \\  \hline \hline
    \multirow{2}{*}{$\mathcal{L}_{\rm{recon}}$} & $\lambda_{\rm{valid}}$       & 1.         \\ \cline{2-3} 
                                      & $\lambda_{\rm{hole}}$        & 6.      \\ \hline
    \multirow{3}{*}{$\mathcal{L}_{\rm{ps}}$}    & $\lambda_{l_1}$          &  1.          \\ \cline{2-3} 
                                      & $\lambda_{\rm{gan}}^{pd}$    &    0.05       \\ \cline{2-3} 
                                      & $\lambda_{\rm{percep}}$      &     0.02      \\ \hline
    \multirow{2}{*}{$\mathcal{L}_{\rm{blend}}$} & $\lambda_{\rm{tv}}$          &    0.02       \\ \cline{2-3} 
                                      & $\lambda_{\rm{gan}}^{blend}$ &    0.02     
    \end{tabular}
    \vspace{-10pt}
\end{table}

\begin{table*}[tb]
    \centering
    \caption{The evaluation results of PatchMatch~\cite{barnes2009patchmatch}, GL~\cite{iizuka2017globally}, PICNet~\cite{zheng2019pluralistic}, Edge~\cite{nazeri2019edgeconnect}, StructureFlow~\cite{ren2019structureflow}, DeepFill~\cite{yu2018free} and CRA~\cite{yi2020contextual} over Places~\cite{zhou2017places} validation set. $\downarrow$ the lower the better, $\uparrow$ the higher the better. We report the inference speed for different methods on a single V100 GPU. $^\dagger$As we cannot obtain the original trained weights of GL, we re-implemented this method according to their paper \cite{iizuka2017globally}.}
    \vspace{-5pt}
    \begin{tabular}{c|c|c|c|c|c|ccccc|c|c}
     & \multicolumn{5}{c|}{random rectangle mask} & \multicolumn{5}{c|}{random irregular mask}  & \multicolumn{2}{c}{inference}                                                                           \\ \cline{2-13}
     & $l_1$ $\downarrow$ & PSNR$\uparrow$ & SSIM$\uparrow$ & TV $\downarrow$ & FID$\downarrow$ & \multicolumn{1}{c|}{$l_1$ $\downarrow$} & \multicolumn{1}{c|}{PSNR$\uparrow$} & \multicolumn{1}{c|}{SSIM$\uparrow$} & \multicolumn{1}{c|}{TV $\downarrow$} & FID$\downarrow$ & FPS & \#param\\ \hline \hline
     PatchMatch     &14.795&15.038&0.819&\textbf{10.93}&11.630& \multicolumn{1}{c|}{11.276} & \multicolumn{1}{c|}{17.387} & \multicolumn{1}{c|}{0.839} & \multicolumn{1}{c|}{\textbf{13.02}} & 10.751 & 11 & -- \\
     GL$^\dagger$ &13.806&15.659&0.821&12.04&10.379& \multicolumn{1}{c|}{10.269} & \multicolumn{1}{c|}{17.403} & \multicolumn{1}{c|}{0.855} & \multicolumn{1}{c|}{13.50} & 8.295 & 24 & 3.6M \\
     PICNet &12.722&16.068&0.801&12.64&9.638 & \multicolumn{1}{c|}{9.477} & \multicolumn{1}{c|}{18.097} & \multicolumn{1}{c|}{0.860} & \multicolumn{1}{c|}{13.35} & 8.097 & 22 & 6.0M\\
     Edge&11.105&16.690&0.858&11.30&8.176& \multicolumn{1}{c|}{9.368} & \multicolumn{1}{c|}{18.249} & \multicolumn{1}{c|}{0.869} & \multicolumn{1}{c|}{13.44} & 8.097 & 20 & 5.3M \\
     StructureFlow&10.915&16.850&0.861&11.18&8.156& \multicolumn{1}{c|}{9.348} & \multicolumn{1}{c|}{18.279} & \multicolumn{1}{c|}{0.872} & \multicolumn{1}{c|}{13.24} & 8.080 & 18 & 10.2M \\
     DeepFill&10.829&16.843&0.859&11.35&8.148& \multicolumn{1}{c|}{9.372} & \multicolumn{1}{c|}{18.230} & \multicolumn{1}{c|}{0.871} & \multicolumn{1}{c|}{13.42} & 8.079 & 45 & 4.1M\\ 
     CRA&10.830&16.839&0.861&11.40&8.150& \multicolumn{1}{c|}{\textbf{9.260}} & \multicolumn{1}{c|}{18.226} & \multicolumn{1}{c|}{0.870} & \multicolumn{1}{c|}{13.33} & 8.071 & 48 & 2.7M\\ \hline 

     Our T-MAD &\textbf{10.334}&\textbf{17.203}&\textbf{0.867}&11.04&\textbf{8.131}& \multicolumn{1}{c|}{\textbf{9.261}} & \multicolumn{1}{c|}{\textbf{18.351}} & \multicolumn{1}{c|}{\textbf{0.873}} & \multicolumn{1}{c|}{13.27} & \textbf{8.058} & 47 & 3.9M
    \end{tabular}
    \label{tab:places}
    \vspace{-10pt}
\end{table*}

\begin{table}[tb]
    \centering
    \caption{The evaluation results of PatchMatch~\cite{barnes2009patchmatch}, PICNet~\cite{zheng2019pluralistic}, Edge~\cite{nazeri2019edgeconnect}, StructureFlow~\cite{ren2019structureflow}, and DeepFill~\cite{yu2018free} over CelebA-HQ and Paris Street-View validation set. $\uparrow$ the higher the better.}
    \vspace{-10pt}
    \begin{tabular}{c|c|c|cc}
     & \multicolumn{2}{c|}{CelebA-HQ} & \multicolumn{2}{c}{Paris Street-View}                                                                             \\ \cline{2-5}
     &  PSNR$\uparrow$ & SSIM$\uparrow$  & \multicolumn{1}{c|}{PSNR$\uparrow$} & SSIM$\uparrow$   \\ \hline \hline
     PatchMatch     &16.014&0.750& \multicolumn{1}{c|}{17.216}& 0.801 \\
     PIC     &25.210&0.852& \multicolumn{1}{c|}{25.620}& 0.834 \\
     Edge     &25.289&0.856& \multicolumn{1}{c|}{25.910}& 0.877 \\
     StructureFlow     &23.644&0.839& \multicolumn{1}{c|}{25.821}& 0.875 \\
     DeepFill     &	25.721 &0.871& \multicolumn{1}{c|}{26.012}& \textbf{0.881} \\ \hline 

     Our T-MAD &\textbf{26.138}&\textbf{0.882}& \multicolumn{1}{c|}{\textbf{26.153}} & 0.880
    \end{tabular}
    \label{tab:celeba-paris}
    \vspace{-20pt}
\end{table}

\begin{figure*}[t]
    \centering
    \includegraphics[width=0.99\linewidth]{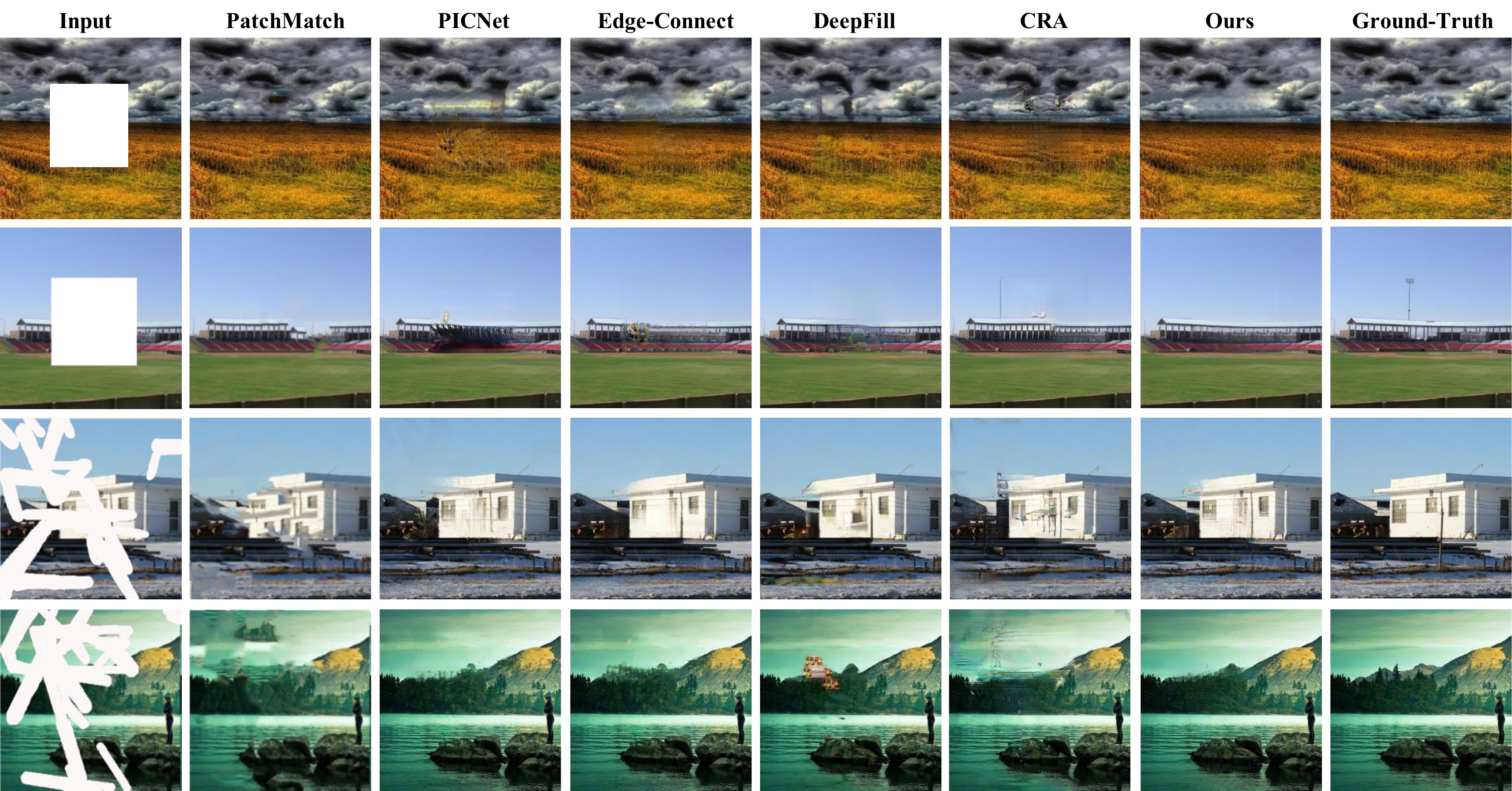}
    \vspace{-5pt}
    \caption{The qualitative comparison with existing models. From left to right: Corrupted input image, results of PatchMatch~\cite{barnes2009patchmatch}, PICNet~\cite{zheng2019pluralistic}, Edge-Connect~\cite{nazeri2019edgeconnect}, DeepFill~\cite{yu2018free}, CRA~\cite{yi2020contextual}, our T-MAD and ground-truth. (\textbf{Best viewed with zoom-in})}
    \label{fig:places-res}
    \vspace{-15pt}
\end{figure*}

To process patches in an efficient way within the existing deep learning framework for training, for each image, a fixed number $\mathbb{N}_{p_s}$ of $p_s$ are extracted and concatenated in an extra dimension. 
Once the number of actual masked patches is smaller than $\mathbb{N}_{p_s}$, we will randomly sample some valid patches from the unmasked region as complementary patches. As for testing, we can merely choose the masked patches to process in the patch synthesis stage.
In our experiments, we keep the patch size $k_p=32$. Because too small patches will cause boundary artifacts while large patches result in heavier training costs for obtaining a high-quality patch synthesis network.
The coarse network is pre-trained for four epochs to obtain a reasonable initialization for the following joint training procedure. 
In the joint training stage, we use four images in each batch and each image has different holes.
Note that PSNet adopts patches as input. The batch size for PSNet is equal to four times the number of hole patches in each image.
Thus, we do not need large image batches to perform adversarial training.
The whole model is trained on four Titan V GPUs for three days. 
In training, we use images of resolution $256 \times 256$ with the largest hole size $128\times 128$ in random positions. 
Cosine restart scheduler is also adopted as our training scheduler.
For each dataset mentioned below, we follow the official partition of the training and validation subsets.
%

As for the patch distribution loss described in Sec.~\ref{sec:psmodule}, we have the freedom of choosing  different $f$ and $g$ for the general formulation like ~\cite{jolicoeur2018relativistic}. In this paper, we apply classical non-saturated GAN loss in~\cite{goodfellow2014generative}. The hyper-parameters of losses are presented in Tab.~\ref{tab:param}.

%
%

\begin{figure*}[t]
    \centering
    \includegraphics[width=0.85\linewidth]{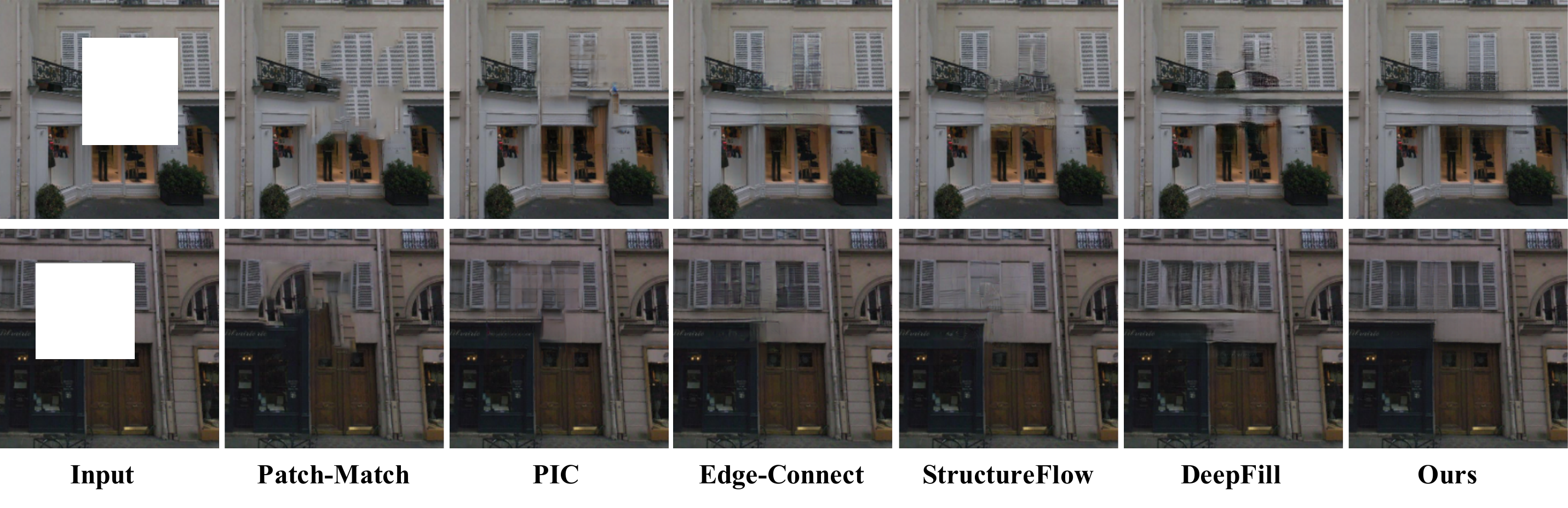}
    \vspace{-10pt}
    \caption{Comparison of our model with PatchMatch~\cite{barnes2009patchmatch}, PIC~\cite{zheng2019pluralistic}, Edge-Connect~\cite{nazeri2019edgeconnect}, StructureFlow~\cite{ren2019structureflow}, and DeepFill~\cite{yu2018free} in Paris Street-View\cite{doersch2012what} validation set. (\textbf{Best viewed with zoom-in})}
    \label{fig:paris}
    \vspace{-10pt}
 \end{figure*}
 
 \begin{figure*}[t]
    \centering
    \includegraphics[width=0.85\linewidth]{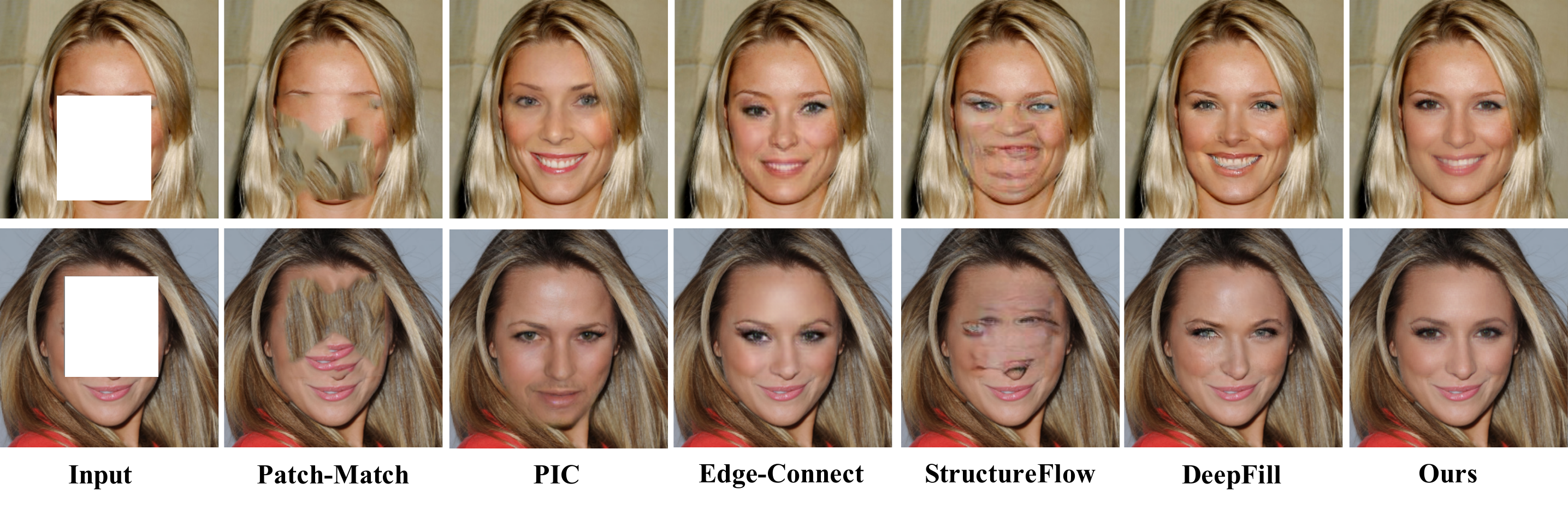}
    \vspace{-10pt}
            \caption{Comparison of our model with PatchMatch~\cite{barnes2009patchmatch}, PIC~\cite{zheng2019pluralistic}, Edge-Connect~\cite{nazeri2019edgeconnect}, StructureFlow~\cite{ren2019structureflow}, and DeepFill~\cite{yu2018free} in CelebA-HQ validation set. (\textbf{Best viewed with zoom-in})}
    \label{fig:celeba}
    \vspace{-10pt}
 \end{figure*}

 \begin{figure}[tb]
    \centering
    \includegraphics[width=\linewidth]{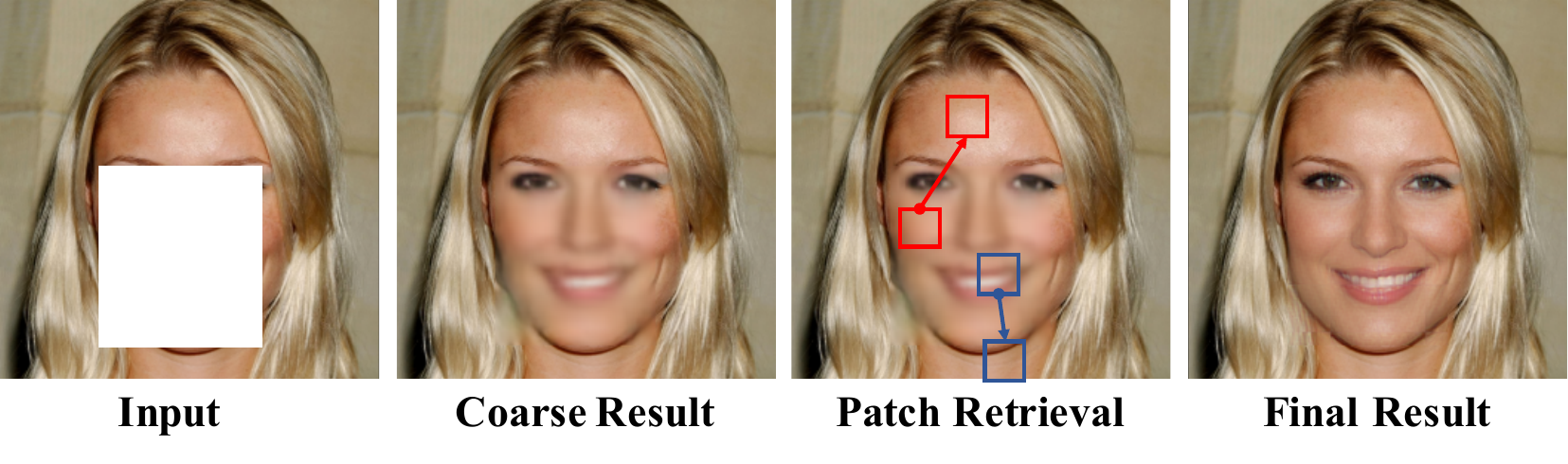}
    \vspace{-15pt}
    \caption{Visualization for the face inpainting procedure with our T-MAD. In the patch retrieval stage, we only show the retrieved patch with the highest score.}
    \label{fig:face}
    \vspace{-15pt}
\end{figure}

\subsection{Main Results}

We evaluate the proposed T-MAD approach on three standard benchmarks Places~\cite{zhou2017places}, CelebA-HQ~\cite{karras2017progressive} and Paris Street-View~\cite{doersch2012what}. 
Among the three datasets, Places is the most challenging benchmark with more than 400 natural scenes of diverse objects and textures. 
CelebA-HQ and Paris Street-View comprise highly structured face and building images, respectively. 
Following the previous study~\cite{yu2018generative}, we use random rectangle masks with the same settings for evaluation. In addition, we also report results on Places with free-form irregular holes~\cite{yu2018free} to further demonstrate the effectiveness of T-MAD for handling various kinds of masks.

\noindent \textbf{Baselines.} We compare our methods with traditional optimization-based methods and contemporary deep inpainting methods:
\begin{itemize}
    \item PatchMatch~\cite{barnes2009patchmatch}: fills in the hole with the most optimal patches from unmasked regions.
    \item GL~\cite{iizuka2017globally}: introduces two discriminators to keep global and local consistency in the results.
    \item PICNet~\cite{zheng2019pluralistic}: introduce random noises for generating diverse results with deep generative network.
    \item Edge~\cite{nazeri2019edgeconnect}: adopts edge information to help reconstruct global structure.
    \item StructureFlow~\cite{ren2019structureflow}: applies the Relative Total Variation (RTV) map~\cite{xu2012structure} guiding the synthesis of global structure.
    \item DeepFill~\cite{yu2018free}: combines gated convolution and contextual attention module for high-quality results. To extract patches from unmasked regions, the contextual attention module adopts softmax function to fuse all of the features from valid regions.
    \item CRA~\cite{yi2020contextual}: improves DeepFill~\cite{yu2018free} with multi-scale contextual attention mechanism using shared attention scores.
\end{itemize}

\noindent\textbf{Quantitative Results.}
Following previous studies, we employ two kinds of metrics to measure distortion and perceptual quality, respectively. 
For distortion measurement, $l_1$ error, Peak Signal-to-Noise Ratio (PSNR), structural similarity index (SSIM), and total variation loss (TV loss) are adopted. 
As for the perceptual measurement, we use Fr\'{e}chet Inception Distance (FID)~\cite{heusel2017gans} to show the Wasserstein-2 distance between two distributions. 
The evaluation results on Places validation set are reported in Tab.~\ref{tab:places}.
Except for GL~\cite{iizuka2017globally}, we directly use the released pre-trained weights and evaluate them in the same testing scheme.
The quantitative results of the baseline methods may be different from their papers, because we apply different testing schemes, e.g., the setting of mask ratios.

The proposed T-MAD achieves competitive results compared with previous works. 
The better performance in PSNR and SSIM demonstrates that the patch-based generation can also contribute effectively to structural recovery. 
The lower FID score suggests our method is more effective in synthesizing realistic texture on such challenging dataset. It is noted that PatchMatch is doing better on TV metric because the method directly copies raw image patches. Although raw image patches bring a lower total validation loss, the overall structure in PatchMatch's results are not satisfactory.
For the inference speed, our T-MAD is even faster than the popular DeepFill methods. Without the time-consuming iterative optimization procedure used in PatchMatch, our texture memory and differentiable retrieval module can work in a more time-efficeint way.



Besides, Tab.~\ref{tab:celeba-paris} provides quantitative comparisons on CelebA-HQ and Paris Street-View dataset. Since these two datasets contain few samples in the validation set, we only report the PSNR and SSIM metrics.
The competitive PSNR and SSIM in Tab.~\ref{tab:celeba-paris} prove the effectiveness of our T-MAD in such specific scenarios that contain highly structured patterns. 
We also observe that the performance of various methods in Places dataset is consistently worse than that in the other datasets, \ie, CelebA-HQ, and Paris StreetView.
A possible reason for this phenomenon is that CeleA-HQ and Paris StreetView only cover homogeneous domains within a single scenario, \ie, faces, or street views.
Nevertheless, Places dataset collects images from more diverse scenes and offers a larger validation set, which is more challenging in the inpainting task.

\noindent\textbf{Qualitative Results.}
Qualitative results are shown in Fig.~\ref{fig:places-res}. It is observed that the traditional patch-based method PatchMatch~\cite{barnes2009patchmatch} fills in the missing region with native texture patches but neglects the global structure. This is in concordance with its lower total variation loss but poorer performance on other metrics in Tab.~\ref{tab:places}.
Existing deep methods~\cite{nazeri2019edgeconnect,yu2018free,yi2020contextual} perform better structural recovery but the generated textures still suffer from artifacts.
On the contrary, our approach generates fine textures faithful to the unmasked regions, while obeying the global structure of the scene (\eg, the highly structured building in the second case). 
In addition, DeepFill~\cite{yu2018free} and CRA~\cite{yi2020contextual} both apply contextual attention mechanism to borrow information from valid regions, which is similar with our texture memory in some extent. However, they use a softmax function to fuse all of the features from unmasked regions and directly generate results at image level, which causes unnatural artifacts.
Thanks to the texture memory and patch synthesis, our T-MAD can easily handle more general cases in such large-scale dataset~\cite{zhou2017places}.

\begin{table}[tb]
    \small
    \centering
    \caption{Ablation study on the number of candidates in the texture pool $\mathbb{N}_c$. `W-Sum' denotes fusing $\{p_c\}$ by simple weighted-sum with correspondance map $\mathcal{S}$. $\downarrow$ the lower the better, $\uparrow$ the higher the better}
    \vspace{-5pt}
    \begin{tabular}{c|c|c|c|c|c}
     & $l_1$ error$\downarrow$ & PSNR$\uparrow$ & SSIM$\uparrow$ & TV$\downarrow$ & FID$\downarrow$ \\ \hline \hline
     $\mathbb{N}_c = 1$ &10.341&17.119&0.852&11.52&8.151\\
     $\mathbb{N}_c = 2$ &10.336&17.130&0.859&11.46&8.147\\
     $\mathbb{N}_c = 4$ &\textbf{10.334}&17.203&\textbf{0.867}&\textbf{11.23}&8.131\\
     $\mathbb{N}_c = 6$ &10.335&\textbf{17.210}&0.864&11.24&\textbf{8.130}\\ \hline
     W-Sum        &10.602&16.872&0.853&11.31&8.602
    \end{tabular}
    \vspace{-10pt}
    \label{tab:nc}
\end{table}
\begin{figure}[tb]
    \includegraphics[width=\linewidth]{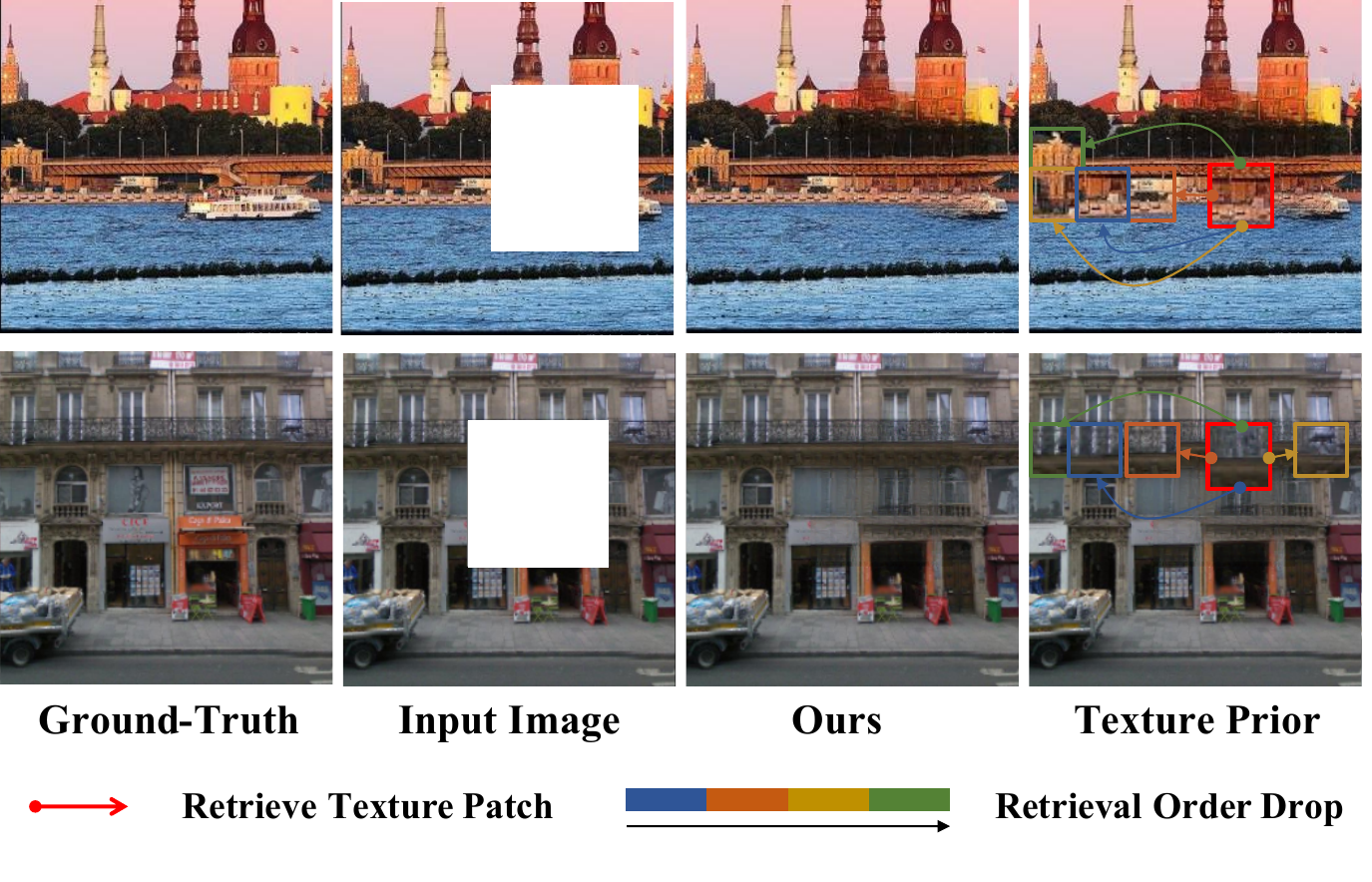}
    \vspace{-10pt}
    \caption{We visualize the retrieval process of texture memory in the last column. The red color represents inpainted patches and other different colors represent the retrieval order of each $p_c^i$. For better visualization, we upsample the cropped patches}
    \label{fig:pool}
    \vspace{-5pt}
\end{figure}

Our method also achieves outstanding perceptual quality in comparison to other methods on highly structured street views (Fig.~\ref{fig:paris}) and faces (Fig.~\ref{fig:celeba}). Some interesting cases are observed.
For corrupted street views, benefiting from the texture memory, our approach learns to borrow similar textures from unmasked appearance of a building, leading to high-quality textures coherent with the original image.

In masked faces, unlike PatchMatch, our model hallucinates reasonable and high-quality content for the missing parts.
Figure \ref{fig:face} presents the intermediate results in our T-MAD. 
The success is partially attributed to the structural guidance offered by the coarse network. The texture memory, interestingly, also functions very well despite the absence of highly identical patches from unmasked regions. The memory, in these examples, offers useful style information of the unmasked regions like complexion and facial texture for the completion. For the mouth patch in Fig.~\ref{fig:face}, PSNet encapsulates the ability of generating novel contents without the help of retrieved patches. Thus, out T-MAD can still achieve high-quality synthesis in such a challenging case.

\subsection{Ablation Study}

\noindent \textbf{Effectiveness of Texture Memory.}
Figure~\ref{fig:pool} presents two actual examples to show which texture patches $p_c^i\in \mathcal{T}_c$ are retrieved from the texture memory $\mathcal{T}$. 
Our approach tends to select patches with similar color or material from the texture memory $\mathcal{T}$.
In the first example of Fig.~\ref{fig:pool}, T-MAD retains a highly faithful texture recurrence with a reasonable semantic structure. It is interesting to see how texture synthesis can benefit from the statistical information provided by the texture memory $\mathcal{T}$.

\begin{figure}[t]
    \includegraphics[width=\linewidth]{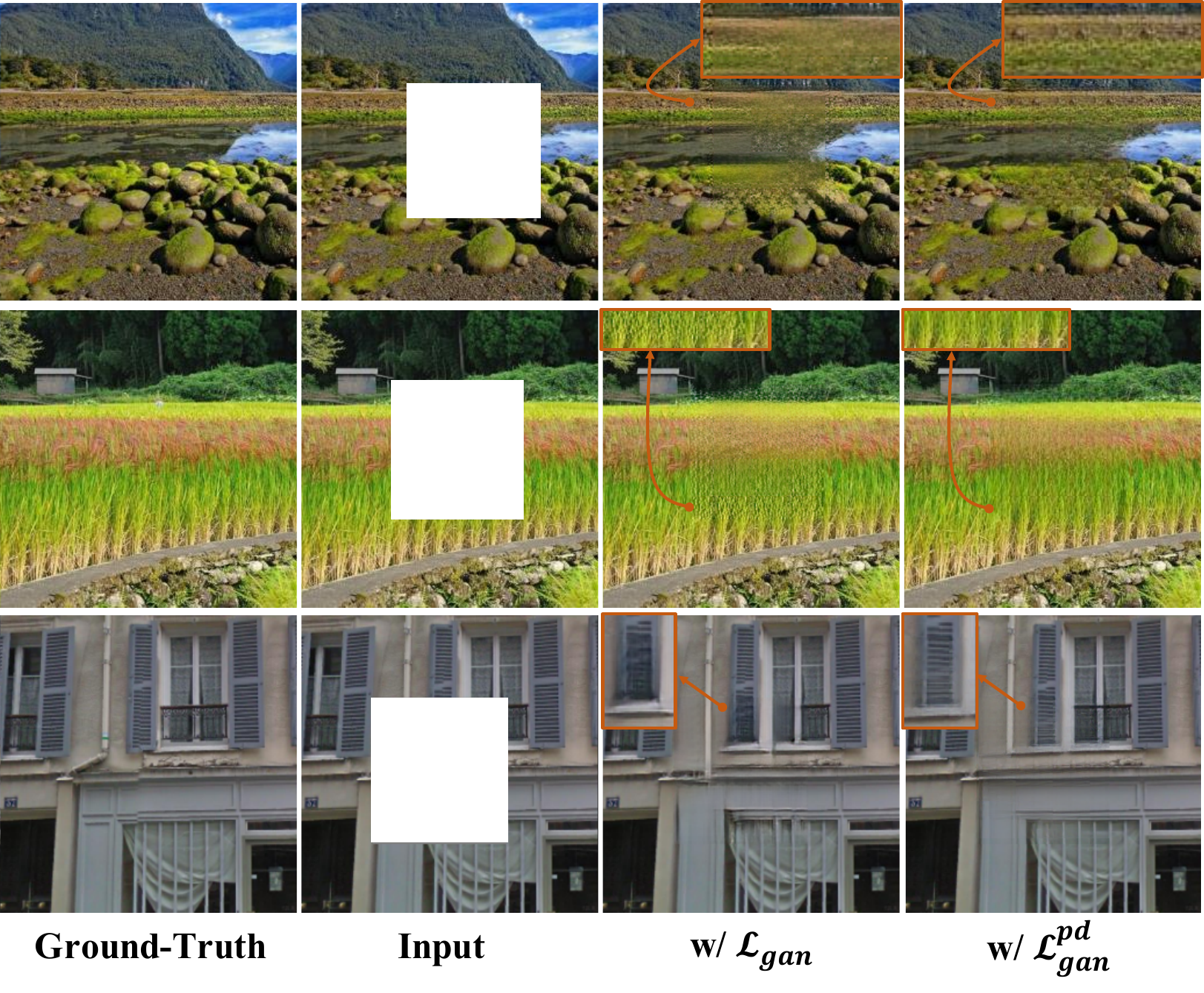}
    \vspace{-10pt}
    \caption{Ablation study on the Patch Distribution loss. Important regions are upsampled for detailed comparison. The first two samples are from Places with diverse scenes, while the last case contains more strctured textures from Paris Street-View. (\textbf{Best viewed with zoom-in})}
    \label{fig:pdloss}
    \vspace{-10pt}
\end{figure}

\begin{table}[tb]
    \caption{Quantitative results for our PSNet with standard GAN loss and our patch distribution loss. $\downarrow$ the lower the better, $\uparrow$ the higher the better}
    \small
    \centering
    \vspace{-5pt}
    \begin{tabular}{c|c|c|c|c}
                   & $l_1$ error $\downarrow$ & PSNR $\uparrow$ & SSIM $\uparrow$ & FID $\downarrow$ \\ \hline \hline
    w/ $\mathcal{L}_{\rm{gan}}$      &10.341&17.200&\textbf{0.869}&8.238\\ \hline
    w/ $\mathcal{L}_{\rm{gan}}^{pd}$ &\textbf{10.334}&\textbf{17.203}&0.867&\textbf{8.131}
    \end{tabular}
    \label{tab:pdloss}
    \vspace{-10pt}
\end{table}

\begin{figure}[t]
    \includegraphics[width=\linewidth]{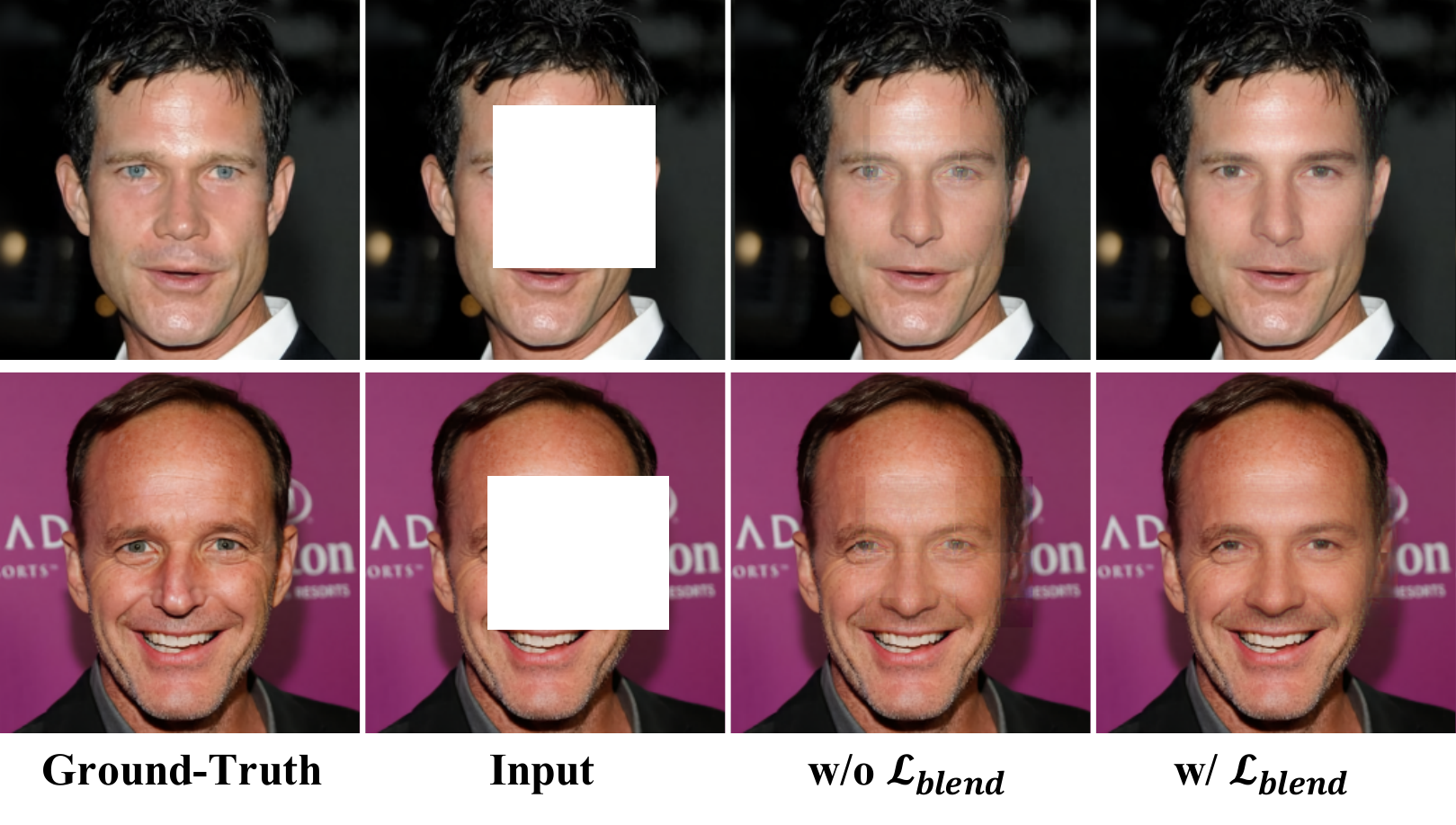}
    \vspace{-10pt}
    \caption{Ablation study on blending loss. (\textbf{Best viewed with zoom-in})}
    \label{fig:blend}
    \vspace{-15pt}
\end{figure}

\begin{figure*}[tb]
    
    \includegraphics[width=0.9\linewidth]{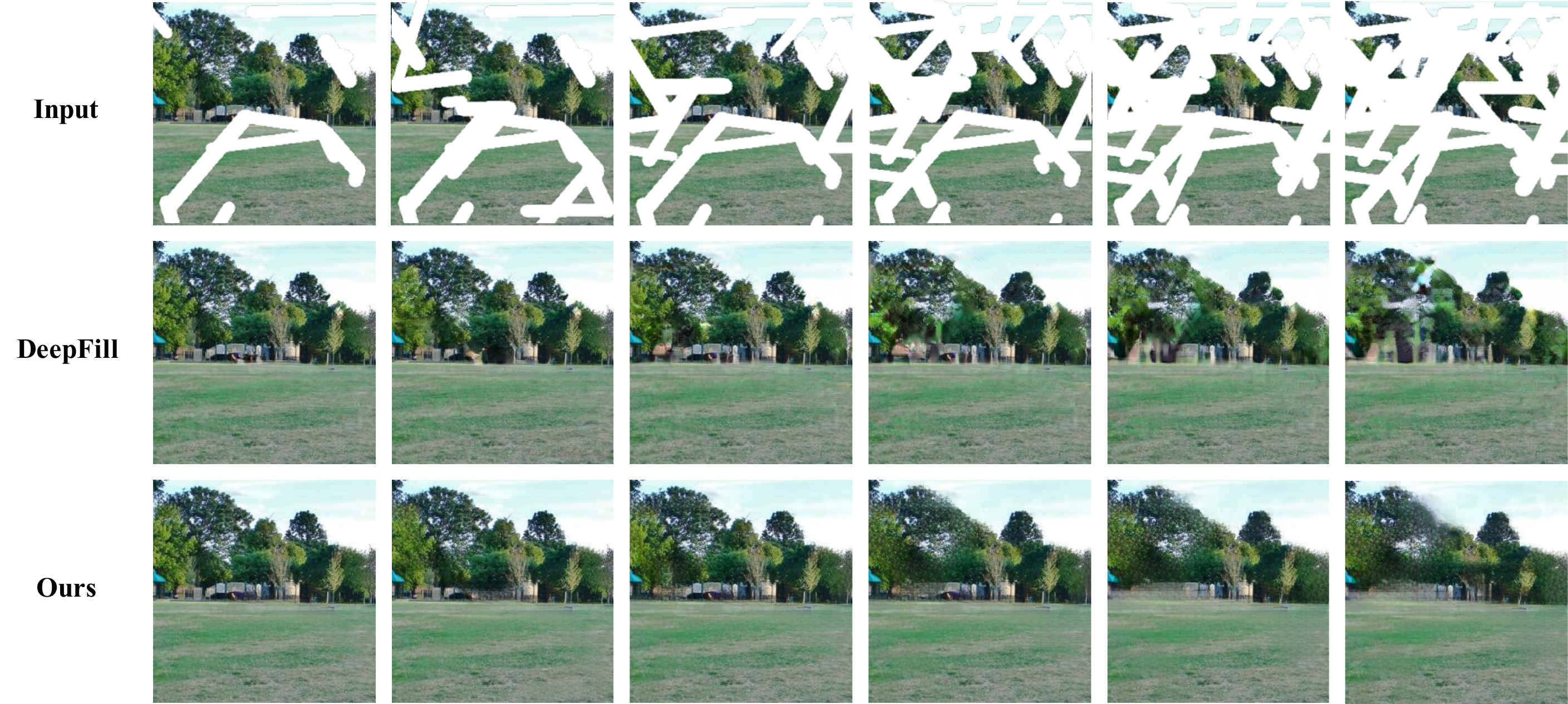}
    \vspace{-5pt}
    \caption{Influence of mask size on inpainting methods. The first row depicts the input with mask showing in white. DeepFill~\cite{yu2018free} (second row) generates undesired artifacts when the area of missing region increases. In contrast, our method (last row) is still capable of synthesizing realistic textures (\textbf{Best viewed with zoom-in})}
    \label{fig:multilevel}
    \vspace{-15pt}
\end{figure*}


The number of candidate texture patches $\mathbb{N}_c$ is a hyper-parameter in the texture retrieval module (Sec.~\ref{sec:pool}). We present the effect of different $\mathbb{N}_c$ in Tab.~\ref{tab:nc}. The `W-Sum' denotes fusing the texture pool with a simple weighted sum given the correspondence map $\mathcal{S}$.
We can clearly see that the weighted sum of texture pool harms the performance in both structure and texture. 
%
%
Selecting fewer texture patches slightly impairs the FID measurement, indicating lower quality of the generated textures. 
%
%
We empirically found that increasing the number of texture patches beyond four does not bring more improvement as more repeated patches do not necessarily bring more statistical information of the unmasked region.

\noindent \textbf{Effectiveness of Patch Distribution Loss.}
An improved adversarial loss at patch level is introduced in Sec.~\ref{sec:psmodule}. 
Figure ~\ref{fig:pdloss} demonstrates the effectiveness of our Patch Distribution loss. We take the model without the relativistic components in Eq.~\eqref{eq:psgan-d} and Eq.~\eqref{eq:psgan-g} as the baseline for this ablation study, denoted by `$w/~\mathcal{L}_{gan}$' in Fig.~\ref{fig:pdloss} and Tab.~\ref{tab:pdloss}. 
We fix other modules and train only PSNet for a fair comparison. 
As shown in Fig.~\ref{fig:pdloss}, even though $\mathcal{T}$ contains useful texture patches, the model without patch distribution loss cannot learn to adopt the texture prior well. The case in Paris Street-View proves the effectiveness of patch distribution loss in synthesizing highly structured textures.
Furthermore, the quantitative results in Tab.~\ref{tab:pdloss} show that $\mathcal{L}_{\rm{gan}}^{pd}$ brings improvement in FID.
%
\begin{table}[tb]
    \caption{Quantitative results for our PSNet as a post-processing module in DeepFill~\cite{yu2018free} framework. $\downarrow$ the lower the better, $\uparrow$ the higher the better}
    \small
    \centering
    \begin{tabular}{c|c|c|c|c}
                   & $l_1$ error $\downarrow$ & PSNR $\uparrow$ & SSIM $\uparrow$ & FID $\downarrow$ \\ \hline \hline
    DeepFill       &10.829&16.843&0.854&8.148\\ \hline
    DeepFill+PSNet &\textbf{10.690}&\textbf{16.967}&\textbf{0.862}&\textbf{8.142}
    \end{tabular}
    \label{tab:post}
    \vspace{-10pt}
\end{table}

\begin{figure}[tb]
    \centering
    \includegraphics[width=\linewidth]{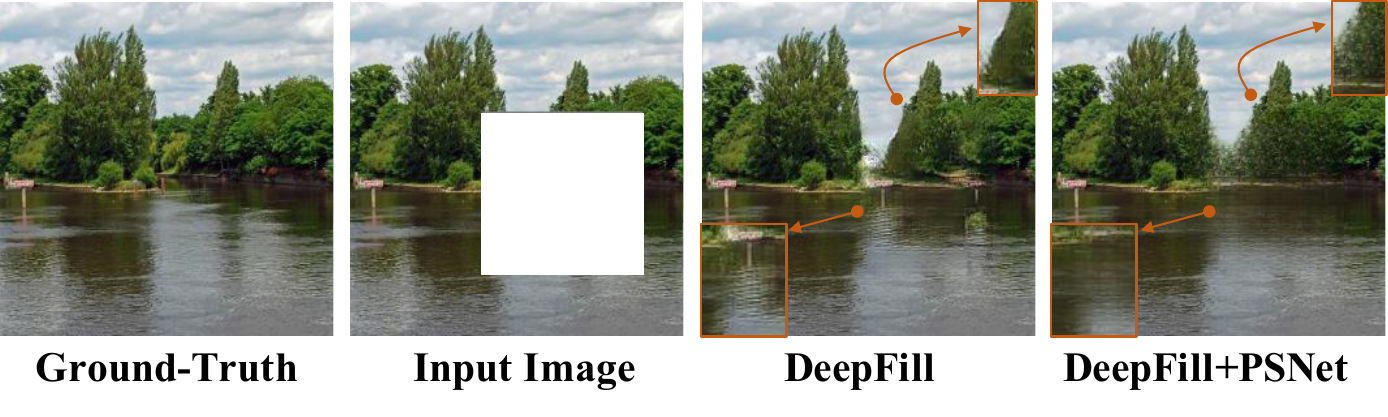}
    \caption{Effectiveness for applying our PSNet as post-processing module for DeepFill. Important regions have been upsampled for detailed comparison}
    \label{fig:post}
    \vspace{-15pt}
\end{figure}

\begin{figure*}[t]
    \centering
    \includegraphics[width=0.9\linewidth]{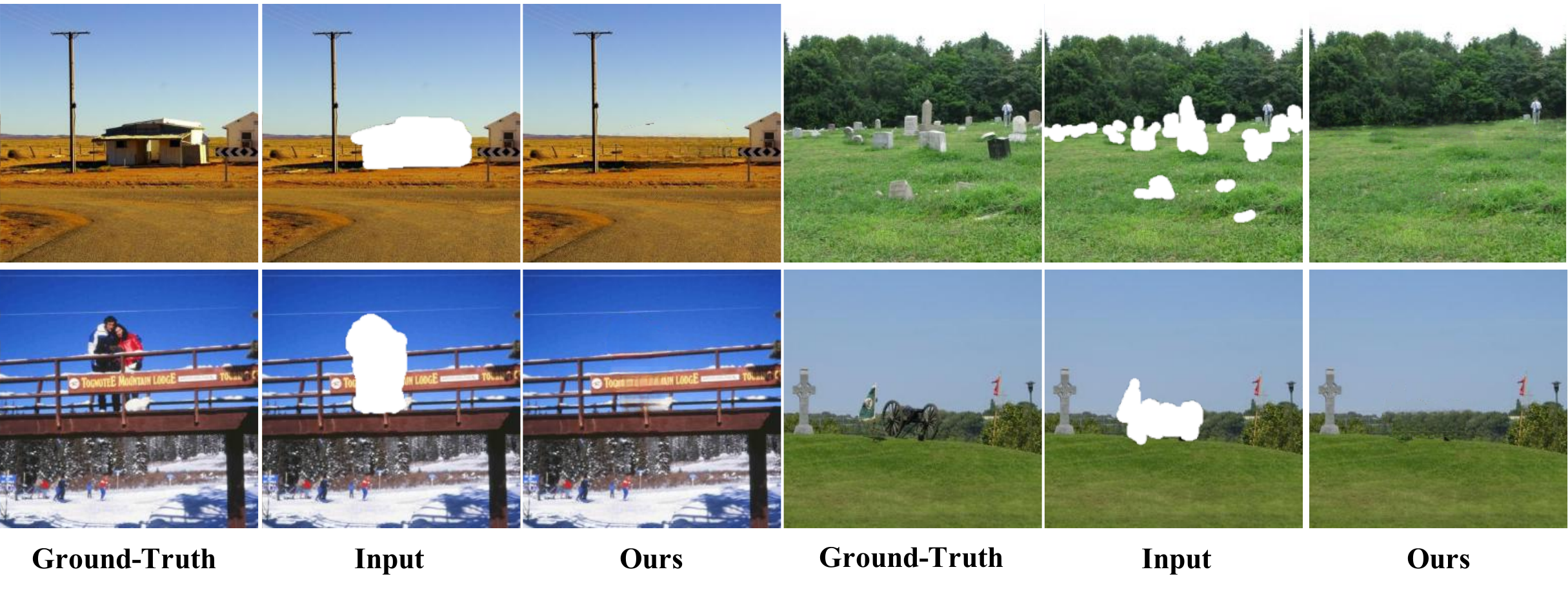}
    \caption{Application for object removal. In each example, from left to right: original image, masked image with white irregular holes and our results (\textbf{Best viewed with zoom-in})}
    \label{fig:object}
    \vspace{-15pt}
\end{figure*}

\begin{figure*}[t]
    \centering
    \includegraphics[width=0.9\linewidth]{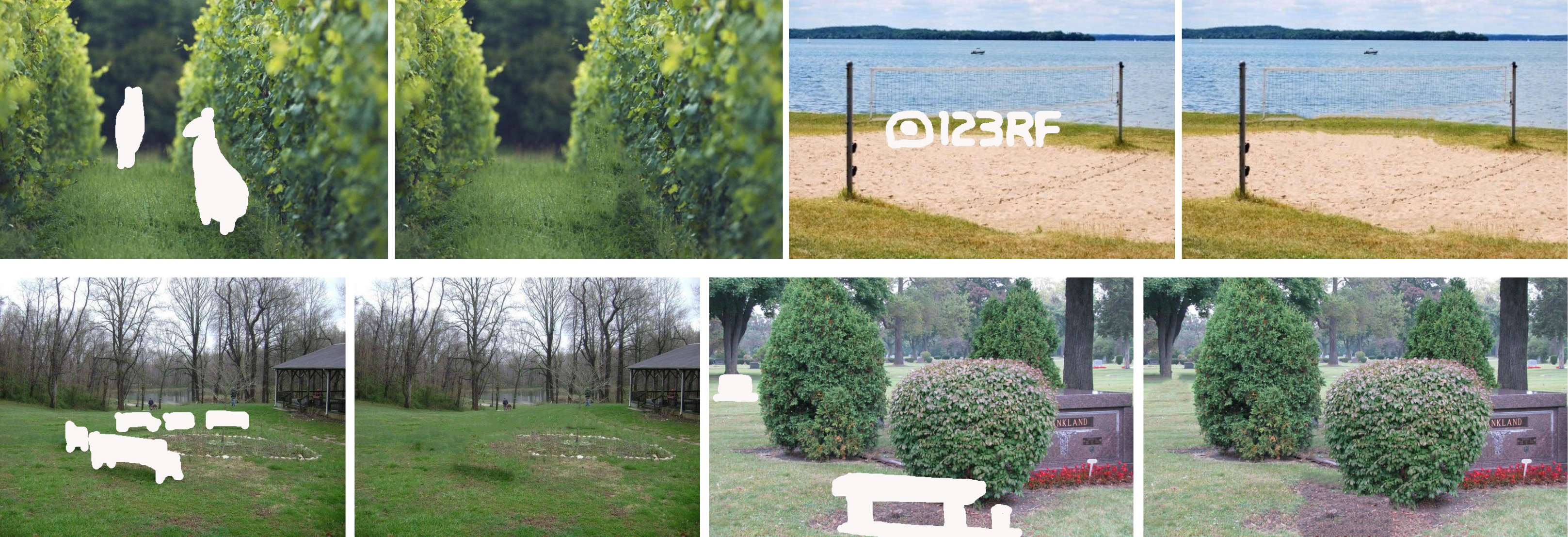}
    \caption{Additional results for high-resolution input. The images shown in this figure are taken from Places val set with their original resolution. (\textbf{Best viewed with zoom-in})}
    \label{fig:supp-hr}
    \vspace{-15pt}
\end{figure*}

\begin{figure*}[t]
    \centering
    \includegraphics[width=0.9\linewidth]{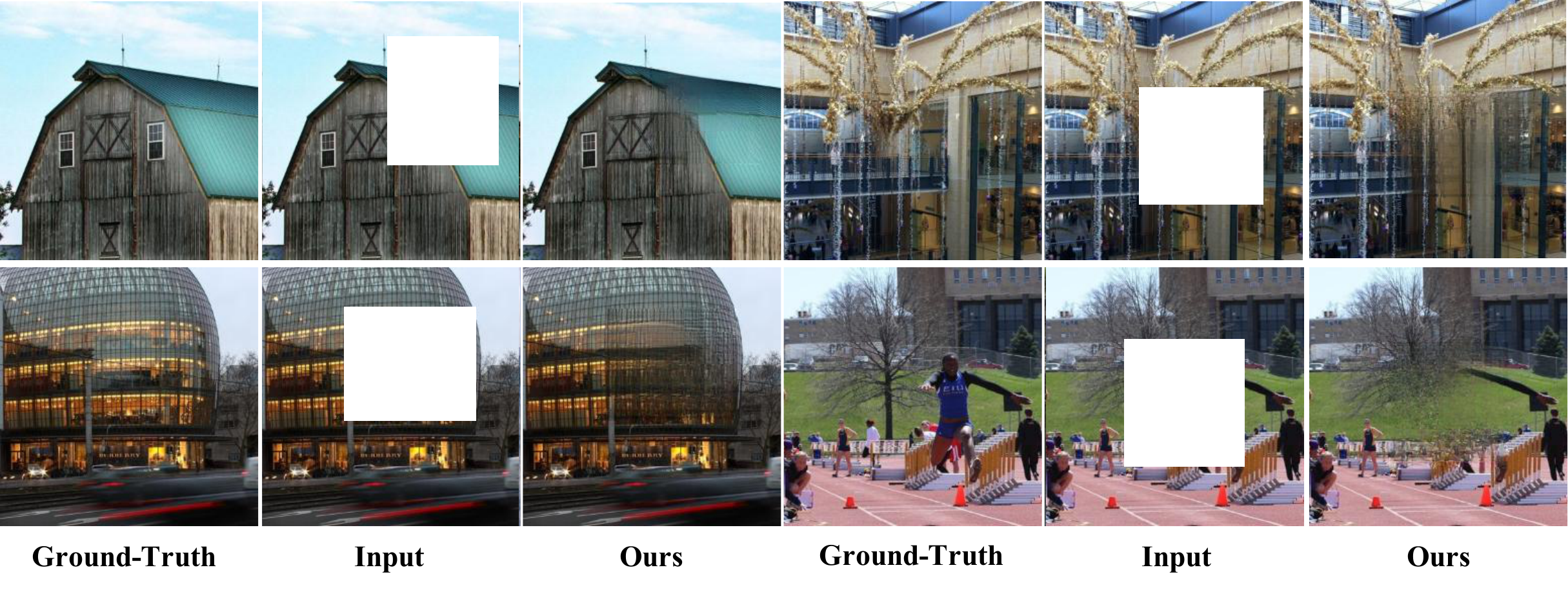}
    \caption{Failure cases of T-MAD}
    \label{fig:fail}
    \vspace{-15pt}
\end{figure*}

\noindent \textbf{Effectiveness of Blending Loss.}
To reduce boundary artifacts and preserve the local consistency between neighboring patches, a blending loss $\mathcal{L}_{blend}$ is adopted in our approach (Eq.~\eqref{eq:blendloss}). 
%
%
We show its importance in our case study on CelebA-HQ.
As shown in Fig.~\ref{fig:blend}, inpainting without the blending loss is susceptible to complex light conditions as well as reflective property of human faces.
%

\noindent \textbf{Influence of Mask Size.}
In Fig.~\ref{fig:multilevel}, we gradually increase the size of input mask to investigate the effects of mask size on inpainting methods. DeepFill~\cite{yu2018free} also adopts memory mechanism with Contextual Attention module. However, as the area of valid region reduces, the method cannot obtain useful information from unmasked regions while our T-MAD can make full use of a few selected patches to synthesize more realistic textures. Furthermore, the interpolation of texture representation in Contextual Attention module also causes undesired artifacts. On the contrary, our method performs better thanks to the more robust patch sampling (see Sec.~\ref{sec:pool}) in texture memory. 

\subsection{Applications}
\noindent \textbf{DeepFill + PSNet.}
As PSNet can also take inpainted images as the input, our patch synthesis module can be easily incorporated into recent inpainting models as a useful post-processing module. 
Taking DeepFill~\cite{yu2018free} method as an example, we adopt the results of DeepFill as input and fine-tune our PSNet with only 10,000 iterations.
As shown in Tab.~\ref{tab:post}, our PSNet improves the quantitative results with just a minor increase in computational cost. The inference time cost introduced by PSNet only accounts for 13\% over the total inference time.
Figure~\ref{fig:post} further demonstrates the effectiveness of PSNet as a post-processing module.
With the help of PSNet, DeepFill can better preserve local consistency in both structure and texture. 
Importantly, the more realistic textures in water and trees verify the importance of applying texture memory in image inpainting.

\noindent \textbf{Object Removal.}
Figure~\ref{fig:object} shows the examples of applying our method to object removal. Users can brush in arbitrary shapes and remove unwanted objects with our approach. The results suggest the generalizability of our approach in dealing with irregular missing regions.

\noindent \textbf{High-Resolution Result.}
To further show the effectiveness of T-MAD with high-resolution input, we present some challenging results in Fig.~\ref{fig:supp-hr}. Even if with high-resolution input ($512\times 768$), our methods can generate high-quality textures matching the original texture distribution.

\subsection{Failure Cases}

Some failure cases are shown in Fig.~\ref{fig:fail} for better understanding the limitation of our method. For most cases, the generated textures are reasonable but the global semantic structure are not fully recovered. It is because the coarse network fails to provide a faithful coarse result due to the complex scenes in the wild. In another failure case that is shown at the bottom-right of Fig.~\ref{fig:fail}, our method fails to hallucinate the whole body of the athlete due to the lack of semantic context.

\section{Conclusion}

We have proposed a novel method that bridges the classic notion of patch-based inpainting and deep learning-based image completion.
Our method uniquely employs a texture memory that comes with an end-to-end trainable texture retrieval module to guide an improved texture generation in a deep inpainting framework. We also introduce a patch distribution loss to enhance texture synthesis at patch level.
Better qualitative and quantitative results against both patch-based and contemporary deep learning-based methods are shown. 
We evisage that the proposed texture memory is not only applicable to image inpainting, but could also benefit other low-level vision tasks such as image super-resolution.

\section*{Acknowledgment}
This study is supported under the RIE2020 Industry Alignment Fund - Industry Collaboration Projects (IAF-ICP) Funding Initiative, as well as cash and in-kind contribution from the industry partner(s). We also thank the supports through the Research Grants Council (RGC) of Hong Kong under ECS Grant No.24206219, GRF Grant No.14204521, CUHK FoE RSFS Grant, SenseTime Collaborative Grant.

\ifCLASSOPTIONcaptionsoff
  \newpage
\fi

\bibliographystyle{IEEEtran}
\bibliography{IEEEabrv,egbib.bib}

\begin{IEEEbiography}[{\includegraphics[width=1in,height=1.25in,clip,keepaspectratio]{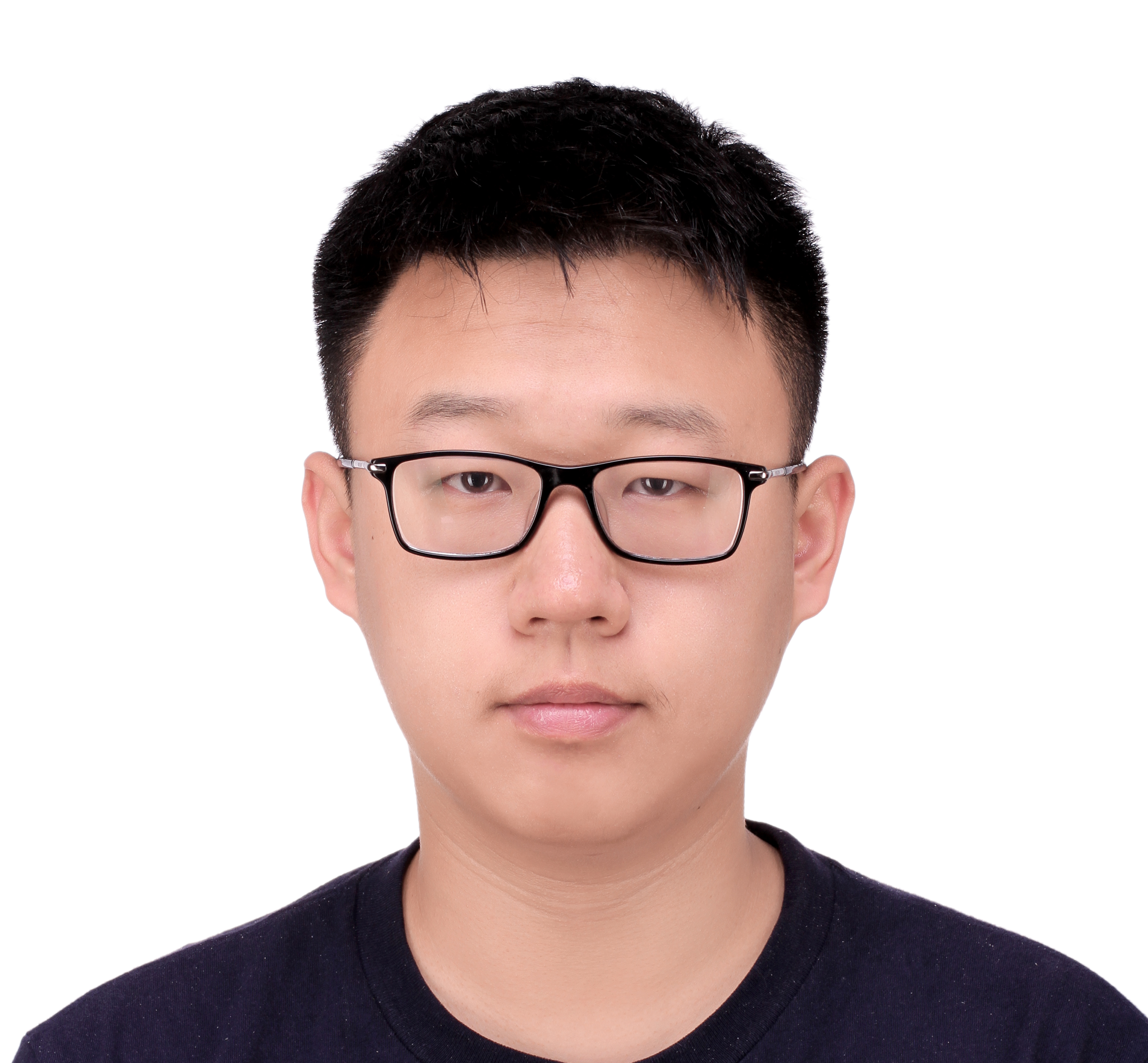}}]{Rui Xu}
  received the BEng degree in electronic engineering from Tsinghua University, in 2018. He is working toward the PhD degree in Multimedia Laboratory (MMLab) at the Chinese University of Hong Kong (CUHK). His research interests include computer vision and deep learning, especially for several topics in low-level vision like inpainting and image generation. He is also a core developer of MMEditing and OpenMMLab project.
\end{IEEEbiography}

\vspace{-15pt}

\begin{IEEEbiography}[{\includegraphics[width=1in,height=1.25in,clip,keepaspectratio]{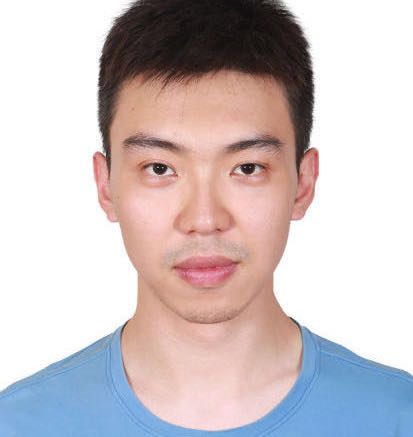}}]{Minghao Guo}
  is a second-year graduate student in Chinese University of Hong Kong. He obtained his BEng major in Automation from Tsinghua University. His research interest lies in computer graphics and computer vision, specifically focusing on data-driven methods, physics-based simulation and inverse problems.
\end{IEEEbiography}

\vspace{-15pt}

\begin{IEEEbiography}[{\includegraphics[width=1in,height=1.25in,clip,keepaspectratio]{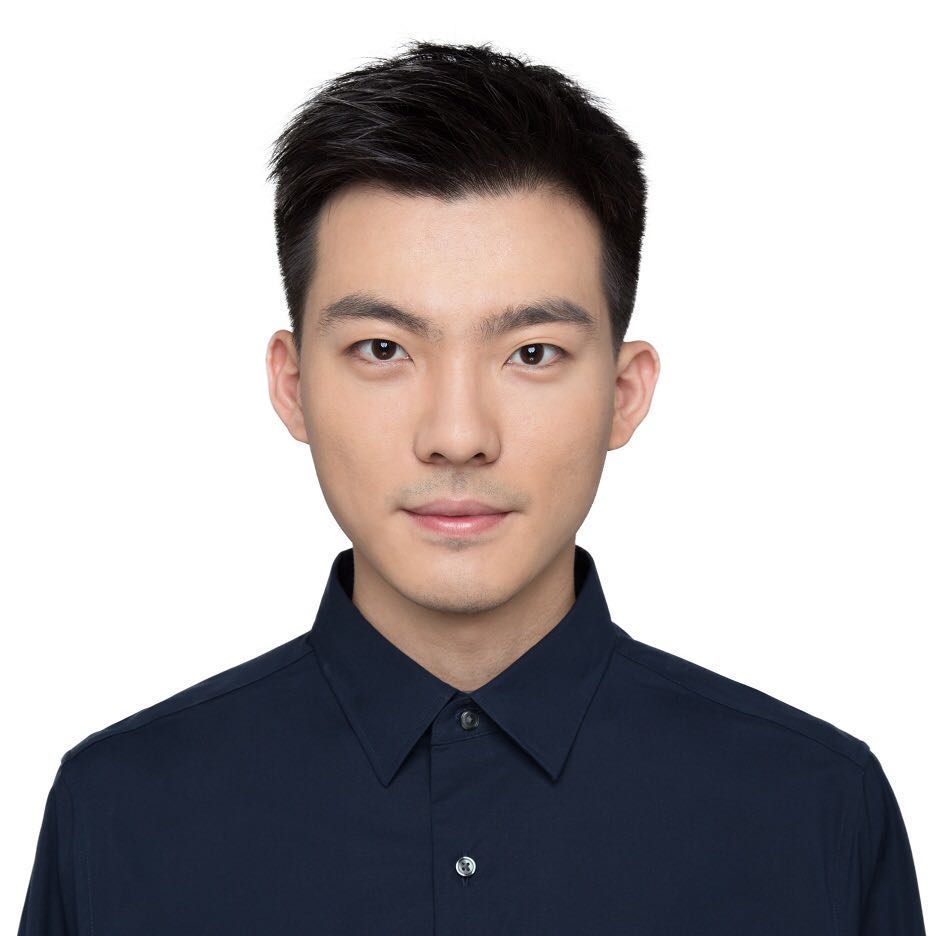}}]{Jiaqi Wang}
  is a fourth-year PhD candidate in Multimedia Laboratory (MMLab) at the Chinese University of Hong Kong. He received the BEng degree from Sun Yat-Sen University in 2017 and began to pursue
his PhD degree funded by Hong Kong PhD Fellowship Scheme (HKPFS) in the same year. His research interests focus on object detection and
instance segmentation. He has 5 papers accepted to CVPR/ICCV/ECCV. He is also a core developer of MMDetection.
His team won the 1st place entry in COCO Object Detection Challenge in 2018 and 2019, respectively.
\end{IEEEbiography}

\vspace{-25pt}

\begin{IEEEbiography}[{\includegraphics[width=1in,height=1.25in,clip,keepaspectratio]{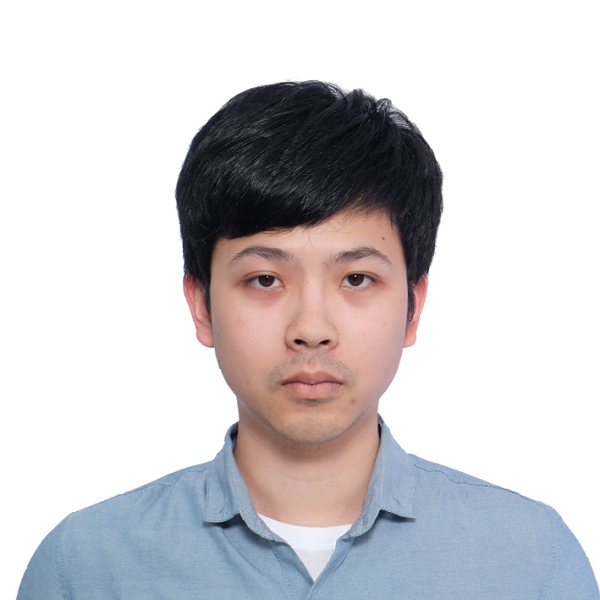}}]{Xiaoxiao Li}
  received his B.E. degree in Department of Computer Science and Technology from Tsinghua University in 2014. He received the Ph.D. degree in 2018 in Information Engineering, Chinese University of Hong Kong (CUHK). He is currently a senior researcher at Gangxing Investment Co. Ltd. His research interests include computer vision and machine learning, especially image segmentation and deep learning.
\end{IEEEbiography}

\vspace{-15pt}

\begin{IEEEbiography}[{\includegraphics[width=1in,height=1.25in,clip,keepaspectratio]{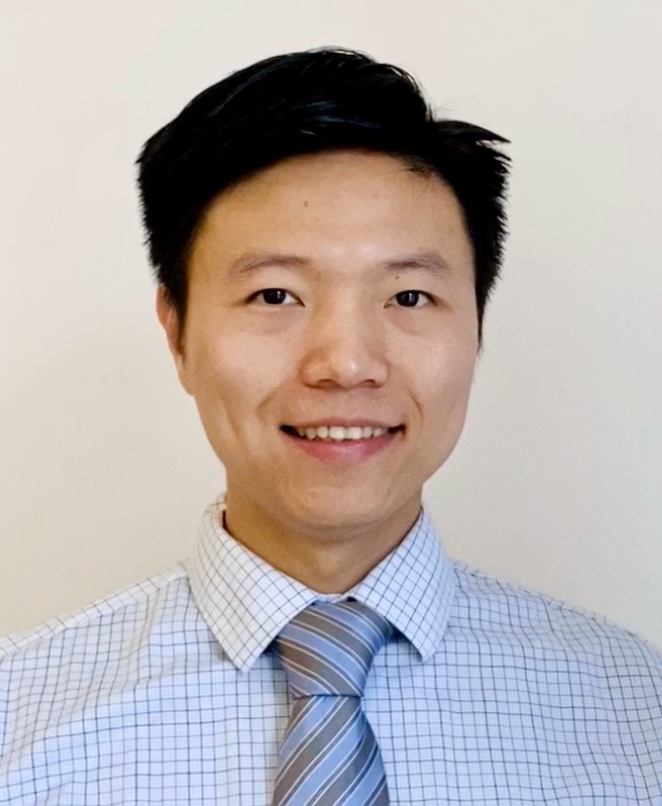}}]{Bolei Zhou}
   received the MPhil degree in information engineering from the Chinese University of Hong Kong and the PhD degree in computer science from Massachusetts Institute of Technology University, in 2018. He is an assistant professor in the Chinese University of Hong Kong. His research interests include computer vision and machine learning, with a focus on enabling machines to sense and reason about the environment through learning more interpretable and structural representations. He is an award recipient of the Facebook Fellowship, the Microsoft Research Asia Fellowship, and the MIT Greater China Fellowship.
\end{IEEEbiography}

\vspace{-15pt}

\begin{IEEEbiography}[{\includegraphics[width=1in,height=1.25in,clip,keepaspectratio]{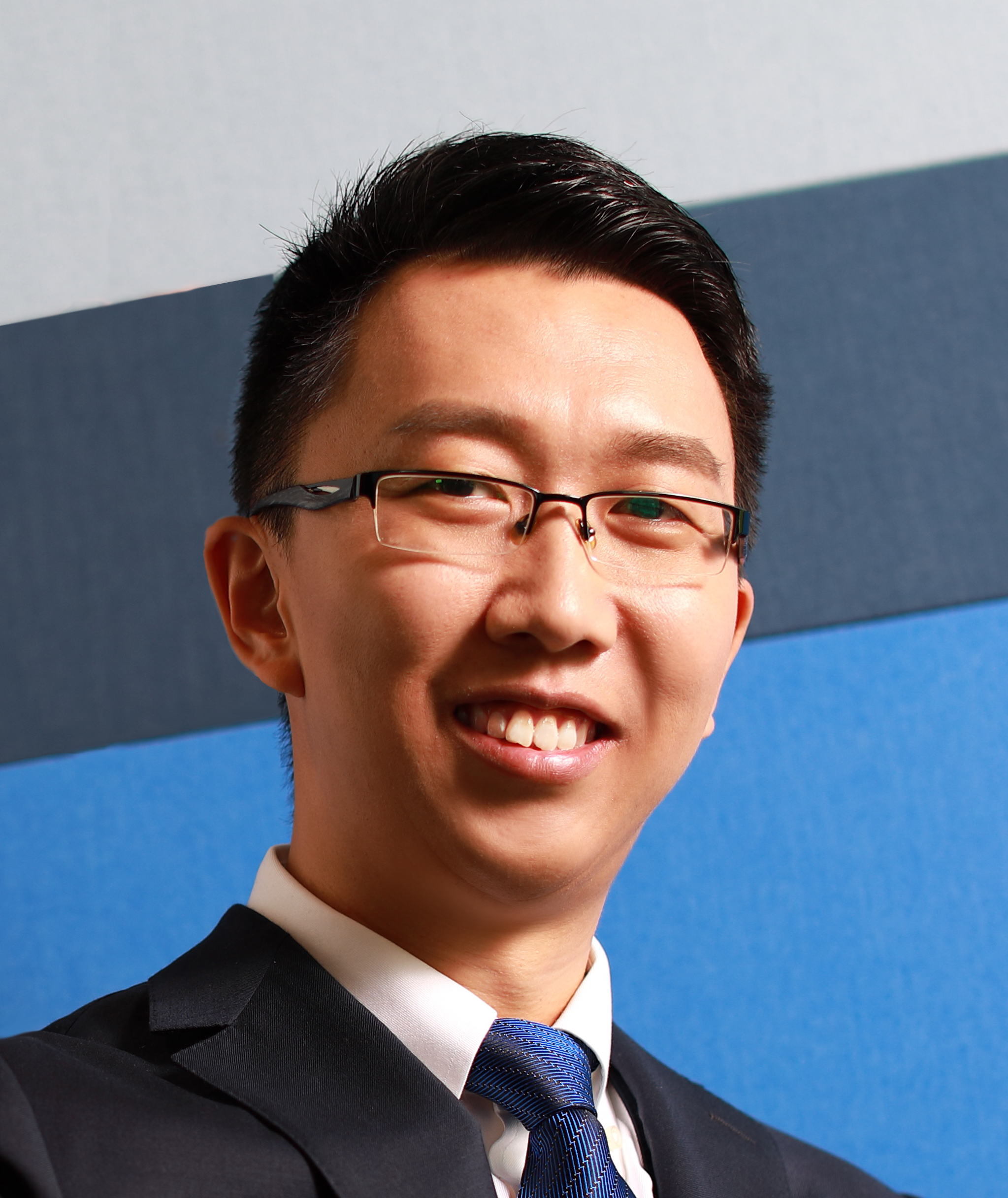}}]{Chen Change Loy} (Senior Member, IEEE) received the PhD degree in computer science from the Queen Mary University of London, in 2010. He is an associate professor with the School of Computer Science and Engineering, Nanyang Technological University. Prior to joining NTU, he served as a research assistant professor with the Department of Information Engineering, The Chinese University of Hong Kong, from 2013 to 2018. His research interests include computer vision and deep learning. He serves as an associate editor of the IEEE Transactions on Pattern Analysis and Machine Intelligence and the International Journal of Computer Vision. He also serves/served as an Area Chair of CVPR 2021, CVPR 2019, ECCV 2018, AAAI 2021 and BMVC 2018-2020. 
\end{IEEEbiography}


\end{document}